\documentclass[pdflatex,bst/sn-mathphys-num]{sn-jnl}
\usepackage{graphicx}
\usepackage{amsmath,amssymb,amsfonts}
\usepackage{amsthm}
\usepackage{mathrsfs}
\usepackage{xcolor}
\usepackage{booktabs}
\usepackage{url}
\usepackage{hyperref}
\usepackage{booktabs}
\usepackage{rotating}
\usepackage{array}
\usepackage{booktabs}
\usepackage{bm}
\usepackage{amstext}
\usepackage{braket}
\usepackage{esint}
\usepackage{colortbl}
\usepackage{float}
\usepackage{flafter}
\usepackage{placeins}
\usepackage{tikz}
\usepackage{tikz-cd}
\usepackage{quantikz}

\definecolor{qGreenFill}{HTML}{E7F4E4}
\definecolor{qGreenDraw}{HTML}{2E7D32}
\definecolor{qBlueFill}{HTML}{E7F0FF}
\definecolor{qBlueDraw}{HTML}{1E5AA8}
\definecolor{qMeasFill}{HTML}{F2F2F2}
\definecolor{qMeasDraw}{HTML}{666666}

\tikzset{
  qboxgreen/.style={draw=qGreenDraw, fill=qGreenFill, rounded corners=2pt, line width=0.4pt,
                    inner xsep=6pt, inner ysep=3pt, align=center},
  qboxblue/.style ={draw=qBlueDraw,  fill=qBlueFill,  rounded corners=2pt, line width=0.4pt,
                    inner xsep=6pt, inner ysep=3pt, align=center},
  meter/.append style={draw=qMeasDraw, fill=qMeasFill, rounded corners=1.5pt, line width=0.35pt}
}

\graphicspath{{./images/}{./figures/}}

\begin{document}
% =========================
% Title + Authors (sn-jnl)
% =========================
\title{Responsible Quantum Machine Learning for Medical Imaging: A Fairness Audit Across Sensitive Subgroups}

\author*[1]{\fnm{Giovanni} \sur{dos Santos Franco}}\email{gs.franco@unesp.br}
\author[1]{\fnm{Felipe} \sur{Mahlow}}\email{f.mahlow@unesp.br}
\author[2]{\fnm{Ellison Fernando} \sur{Cardoso}}\email{Ellison.Cardoso@einstein.br}
\author*[3]{\fnm{José Adenaldo} \sur{Santos Bittencourt}}\email{jose.bittencourt@einstein.br}
\author*[3,1]{\fnm{Felipe} \sur{Fanchini}}\email{felipe.fanchini@einstein.br}

\affil[1]{\orgdiv{Faculty of Sciences}, \orgname{São Paulo State University (Unesp)}, \orgaddress{\city{Bauru}, \country{Brazil}}}
\affil[2]{\orgdiv{Laboratory of Magnetic Resonance in Neuroradiology (LIM-44)}, \orgname{University of São Paulo Faculty of Medicine Clinics Hospital}, \orgaddress{\city{São Paulo}, \country{Brazil}}}
\affil[3]{\orgname{Hospital Israelita Albert Einstein}, \orgaddress{\city{São Paulo}, \country{Brazil}}}

\abstract{
Quantum machine learning (QML) has been increasingly explored for medical classification tasks, yet its behaviour under subgroup-aware fairness criteria remains insufficiently studied. This work presents an empirical fairness audit of classical and quantum classifiers for medical image classification, starting from the premise that QML should not be assumed to be inherently fair. Two healthcare scenarios are evaluated: dermatological lesion classification on DDI, using Fitzpatrick skin type as the sensitive attribute, and chest X-ray classification on CheXpert-derived data, using age group as the sensitive attribute. Using ResNet-18 embeddings as a common feature space, we train classical models (SVM and MLP) and quantum models (QSVM and QCNN), and evaluate them under a baseline setting and three fairness-mitigation settings: differentiable Equalized Odds (EO) regularisation, Gradient Reversal Learning (GRL), and their combination. State-of-the-art foundation models (PanDerm for DDI and EVA-X for CheXpert) are included as external references. The results suggest that quantum classifiers may achieve competitive aggregate performance while still exhibiting subgroup disparities. Because a reduction in the subgroup AUC gap can arise either from improvement of the worst-performing group or from degradation of the best-performing one, each mitigation setting is additionally assessed against a pre-specified non-inferiority margin relative to its own baseline. Under this criterion, several configurations that achieve the smallest absolute disparity---most notably QCNN with GRL on DDI---fail to qualify as clinically acceptable, because the disparity reduction is accompanied by a loss of global discrimination. Among the models trained in this study, QSVM with GRL+EO performed best on DDI and approached the PanDerm reference in the performance--fairness trade-off space, whereas on CheXpert QCNN with GRL achieved the highest aggregate performance among the trained models but increased subgroup disparity relative to its own baseline, and the EVA-X reference remained stronger overall. Overall, the findings indicate that fairness effects are model- and dataset-dependent, and that QML workflows require empirical subgroup-aware evaluation rather than assumptions of inherent fairness. All quantum models are evaluated in classical simulation; the study is therefore an audit of hybrid classical--quantum pipelines and does not claim a quantum advantage in predictive performance, fairness, or computational efficiency.
}

\maketitle

\section{Introduction}\label{sec:introduction}

Artificial intelligence (AI) has become increasingly relevant in medical image analysis, supporting tasks such as dermatological lesion classification, radiological screening, and clinical decision support \cite{Esteva2017Dermatology,Irvin2019CheXpert,SeyyedKalantari2021UnderdiagnosisBias}. Despite strong aggregate performance, medical AI systems may exhibit unequal behaviour across demographic and clinically relevant subgroups, including differences associated with age, sex, ethnicity, and skin tone \cite{Obermeyer2019DissectingBias,Larrazabal2020GenderImbalance}, mirroring broader evidence that AI systems can reproduce and amplify gender and ethnic stereotypes \cite{Franco2026BiasBeyondBorders}. In dermatology, the Diverse Dermatology Images (DDI) dataset has highlighted limitations of AI models across Fitzpatrick skin type groups, while in chest radiography, fairness gaps and underdiagnosis bias have been reported in large-scale datasets such as CheXpert, MIMIC-CXR, and ChestX-ray14 \cite{Daneshjou2022DDI,SeyyedKalantari2020CheXclusion,SeyyedKalantari2021UnderdiagnosisBias}. These findings indicate that medical AI models should not be evaluated only through aggregate performance metrics; instead, it is necessary to analyse how performance varies across clinically and demographically relevant subgroups \cite{Hardt2016Equality,Barocas2023FairnessML,Mehrabi2021Survey}.

At the same time, quantum machine learning (QML) has emerged as a promising research direction at the intersection of quantum computing and machine learning \cite{Biamonte2017QuantumML}. Quantum support vector machines (QSVMs), variational quantum classifiers, and quantum convolutional neural networks (QCNNs) have been investigated as possible classifiers for high-dimensional data representations and hybrid classical--quantum workflows \cite{Havlicek2019Supervised,Cong2019QCNN,Schuld2021KernelMethods}. However, most practical QML studies still focus primarily on predictive performance, compactness, or near-term feasibility. If QML models are to be considered for healthcare-oriented applications, it may be important to evaluate them not only in terms of predictive performance, but also under responsible AI criteria, including fairness across sensitive groups \cite{Barocas2023FairnessML,Mehrabi2021Survey}.

An important caveat applies to how fairness gains should be read in a medical context. A subgroup-disparity metric such as the group-wise AUC gap can decrease for two mechanistically distinct reasons: because the worst-performing group improves, or because the best-performing group deteriorates~\cite{Mittelstadt2024LevellingDown,Zietlow2022LevelingDown}. The two cases are indistinguishable at the level of the disparity metric alone, but they are not clinically equivalent: only the first represents a benefit to patients, whereas the second achieves statistical equality by degrading care for a subgroup that was previously well served~\cite{Mittelstadt2024LevellingDown,Petersen2023EqualPerformance}. In the limiting case, a model that performs uniformly poorly across all groups would be assigned a disparity of zero~\cite{Zietlow2022LevelingDown}. For this reason, the present work does not treat a reduction in the AUC gap as intrinsically desirable. Disparity reductions are instead interpreted jointly with the change in predictive performance relative to the corresponding unmitigated baseline, under an explicit non-inferiority criterion defined in Section~\ref{subsec:evaluation_metrics}, in line with recommendations for fairness evaluation in medical imaging~\cite{RicciLara2022FairnessImaging,Ueda2024FairnessHealthcare}.

This work presents an empirical fairness audit of classical and quantum machine learning models for medical image classification. The objective is to investigate whether QML classifiers exhibit measurable subgroup disparities and how these disparities compare with classical baselines under related experimental conditions. Two healthcare scenarios are considered: dermatological image classification using DDI, with Fitzpatrick skin type as the sensitive attribute, and chest X-ray classification using CheXpert-derived data, with age group as the sensitive attribute \cite{Daneshjou2022DDI,Irvin2019CheXpert}. In both cases, the images are first processed using the standard ResNet-18 preprocessing pipeline and then encoded with a ResNet-18 model pretrained on ImageNet. The resulting embeddings are used as input representations for the downstream classifiers. Classical baselines, including support vector machines and multilayer perceptrons, are compared with quantum models, including QSVMs and QCNNs. In addition, PanDerm and EVA-X are included in the results as state-of-the-art reference baselines for the dermatology and chest X-ray settings, respectively \cite{Yan2025PanDerm,Yao2025EVAX}.

In addition to auditing baseline behaviour, this study evaluates whether standard fairness mitigation strategies can be adapted to QML workflows. Equalized Odds (EO) is implemented as a differentiable training regularisation term rather than as a threshold-based adjustment. During training, the predicted probabilities are separated by true class and sensitive group, and the model is penalised when the group-wise mean predicted probabilities differ within the same class. This encourages the model to reduce group-dependent differences that are related to true positive and false positive behaviour while preserving end-to-end optimisation. Gradient Reversal Learning (GRL), in turn, is used as an adversarial representation-learning mechanism: during training, a sensitive-attribute predictor is encouraged to identify group information, while the feature representation is simultaneously encouraged to make this prediction difficult \cite{Ganin2016DomainAdversarial}. In this work, GRL is investigated as a fairness-aware mechanism for QML-based classifiers, and it is also combined with the differentiable EO regularisation term. To the best of our knowledge, this is one of the first empirical studies to apply GRL for fairness mitigation in QML-based medical image classification, and also one of the first to evaluate a combined GRL and differentiable EO strategy in QML workflows.

The study is guided by the following research questions:

\begin{enumerate}
    \item Do quantum machine learning classifiers exhibit demographic performance disparities in medical image classification tasks?

    \item How do these disparities compare with those observed in classical baselines trained on related feature representations?

    \item How do fairness mitigation strategies affect QML models in terms of predictive performance and subgroup disparity?

    \item When a mitigation strategy reduces subgroup disparity, does it do so while preserving global predictive performance within a pre-specified non-inferiority margin relative to the unmitigated baseline?
\end{enumerate}

To address these questions, the models are evaluated using predictive metrics, including AUC, balanced accuracy, and F1-score, as well as demographic parity gap and AUC gap \cite{Hardt2016Equality,Barocas2023FairnessML,Mehrabi2021Survey}. The evaluation compares baseline models, EO-mitigated models, GRL-based models, and the combined GRL+EO mitigation strategy, allowing the analysis of both predictive performance and subgroup disparity under different fairness-aware configurations.

The main contributions of this work are summarised as follows:

\begin{enumerate}
    \item \textbf{Empirical fairness audit of QML in medical imaging.}
    This work provides, to the best of our knowledge, one of the first empirical audits of demographic bias in QML models for medical image classification, complementing previous conceptual discussions on quantum fair machine learning \cite{Perrier2021QuantumFairML,Guan2022VerifyingFairnessQML}.

    \item \textbf{Empirical use of GRL for fairness mitigation in QML.}
    The study investigates Gradient Reversal Learning as a fairness-aware adversarial mechanism for QML workflows, aiming to reduce the information about sensitive attributes encoded in learned representations \cite{Ganin2016DomainAdversarial}.

    \item \textbf{Empirical evaluation of GRL combined with differentiable Equalized Odds in QML.}
    The work evaluates a combined GRL+EO strategy, integrating adversarial representation learning with a differentiable Equalized Odds regularisation term for QML-based classifiers \cite{Hardt2016Equality,Ganin2016DomainAdversarial}.

    \item \textbf{Comparison between classical and quantum classifiers.}
    The study compares classical and quantum classifiers under a common fairness evaluation framework, allowing the analysis of whether QML models reproduce, amplify, or reduce subgroup disparities observed in classical pipelines \cite{Biamonte2017QuantumML,Havlicek2019Supervised,Cong2019QCNN}.

    \item \textbf{A non-inferiority criterion for reading performance--fairness trade-offs.}
    The work introduces an explicit, pre-specified acceptability criterion under which a mitigation strategy is considered favourable only if it reduces subgroup disparity \emph{and} does not degrade global predictive performance beyond a defined margin relative to its own unmitigated baseline. Applying this criterion to the present results separates configurations whose apparent fairness gain is accompanied by a loss of discriminative capacity from those in which it is not~\cite{Mittelstadt2024LevellingDown,Piaggio2012CONSORT}.

\end{enumerate}

Overall, this paper argues that QML should not be treated as inherently fairness-neutral or inherently fairness-improving. Quantum models may offer different computational structures and representation mechanisms, but they can still be affected by biased data distributions, subgroup imbalance, and sensitive information encoded in feature representations. Therefore, fairness auditing should become a standard part of QML evaluation in healthcare, alongside predictive performance, circuit design, computational cost, and hardware feasibility.

\section{Methods}\label{sec:materials_methods}

\subsection{Study design}\label{subsec:study_design}

This study was designed as an empirical fairness audit of classical and quantum machine learning models for medical image classification. The experimental workflow consisted of five main stages: dataset preparation, image embedding extraction, classifier training, fairness mitigation, and subgroup-aware evaluation. Two medical imaging scenarios were considered. The first scenario used dermatological images from the Diverse Dermatology Images (DDI) dataset, with Fitzpatrick skin type groups as the sensitive attribute \cite{Daneshjou2022DDI}. The second scenario used CheXpert-derived chest radiographs, with age group as the sensitive attribute and pleural effusion as the binary target condition \cite{Irvin2019CheXpert}.

For both datasets, the images were first encoded into fixed-dimensional classical embeddings using a ResNet-18 model pretrained on ImageNet. These embeddings were used as input representations for the downstream classical and quantum classifiers. This design allowed the comparison between classical and quantum models to be performed over a common feature space.

The study compared classical baselines, including support vector machines (SVMs) and multilayer perceptrons (MLPs), with quantum models, including quantum support vector machines (QSVMs) and quantum convolutional neural networks (QCNNs) \cite{Havlicek2019Supervised,Cong2019QCNN,Schuld2021KernelMethods}. In addition, PanDerm and EVA-X were included in the results as state-of-the-art reference baselines for the dermatology and chest X-ray settings, respectively \cite{Yan2025PanDerm,Yao2025EVAX}.

For each applicable classifier, four evaluation settings were considered: baseline classification without fairness intervention, differentiable Equalized Odds (EO) regularisation, Gradient Reversal Learning (GRL), and the combined GRL+EO strategy \cite{Hardt2016Equality,Ganin2016DomainAdversarial}. For each experimental run, the data were split into training, validation, and test sets using a 60/20/20 proportion. The training set was used to fit the classifiers and compute the differentiable mitigation losses, the validation set was used for model selection and threshold calibration when applicable, and the test set was reserved exclusively for the final evaluation reported in Section~\ref{sec:results}.

The experiments were repeated using five random seeds,

\begin{equation}
\mathcal{S} = \{13, 42, 123, 777, 2024\},
\end{equation}

and the reported deviations correspond to the standard deviation across these independent runs. Thus, the uncertainty values reported in the tables reflect variability across seeds, including the effects of data splitting, model initialisation, and training procedures when applicable. They are not resampling-based confidence intervals for the metrics themselves~\cite{Efron1993Bootstrap}, and they are not paired across models~\cite{DeLong1988}. Consequently, they support statements about the stability of a configuration across repetitions, but they do not by themselves license claims of superiority or equivalence between two configurations whose intervals overlap~\cite{AltmanBland1995Absence}. This point is revisited in Sections~\ref{subsec:global_performance} and~\ref{sec:limitations_future_work}.

\begin{figure}[htbp]
    \centering
    \includegraphics[width=\textwidth]{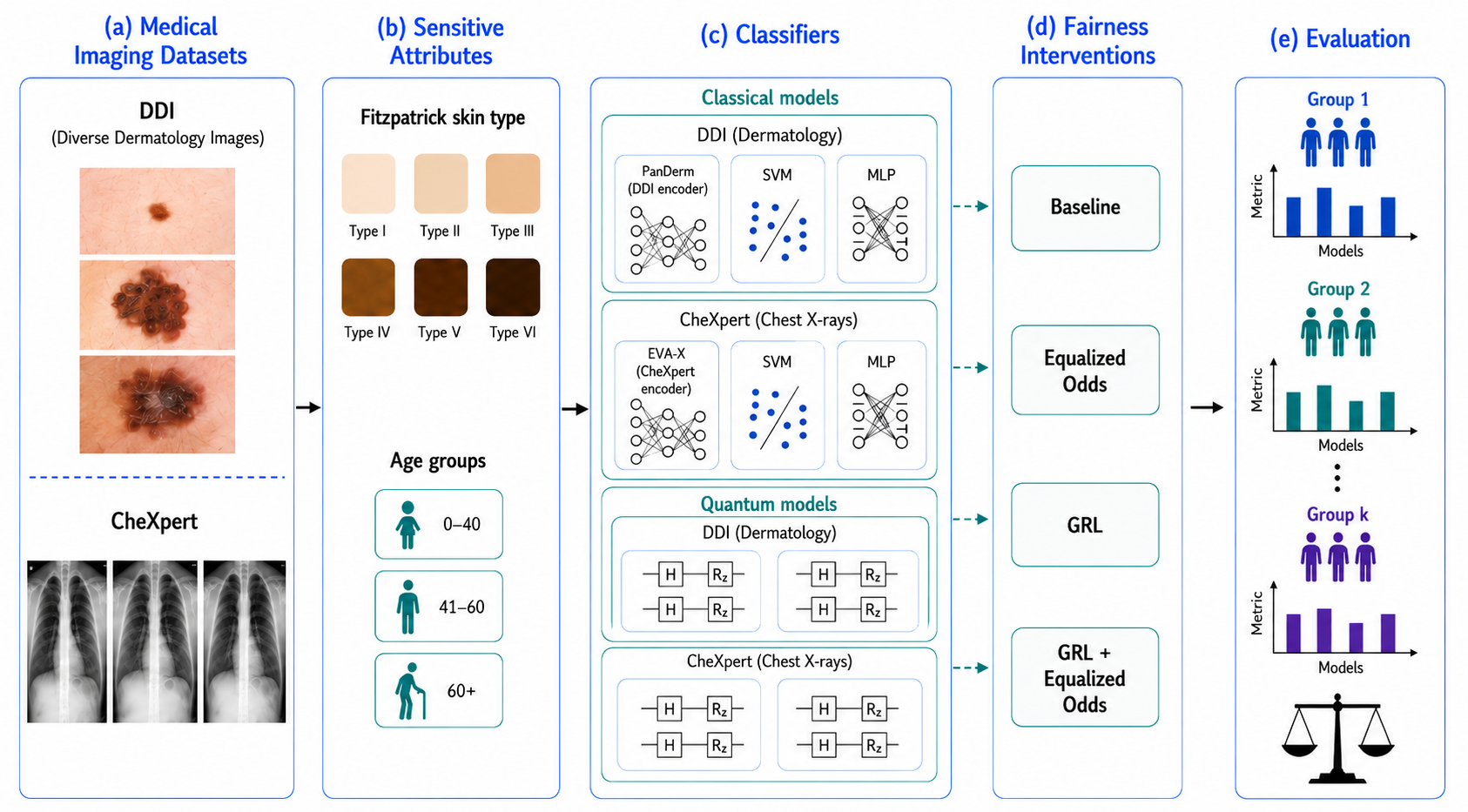}
    \caption{
    \textbf{Bias auditing pipeline for classical and quantum machine learning in medical imaging.}
    The workflow starts from two medical imaging datasets: DDI for dermatology and CheXpert for chest radiography. Sensitive attributes are defined using Fitzpatrick skin type groups for DDI and age groups for CheXpert. Images are encoded using a ResNet-18 model pretrained on ImageNet, and the resulting embeddings are used as inputs to classical and quantum classifiers. The models are evaluated under baseline, EO, GRL, and GRL+EO settings, followed by subgroup-aware analysis using predictive and fairness metrics.
    }
    \label{fig:pipeline_bias}
\end{figure}

The complete experimental pipeline is summarised in Figure~\ref{fig:pipeline_bias}. The figure illustrates the flow from medical imaging datasets and sensitive attributes to embedding extraction, classical and quantum classifiers, fairness interventions, and subgroup-aware evaluation.

%\subsection{Datasets}\label{subsec:datasets}

%This study considered two medical imaging datasets selected to represent distinct clinical domains and sensitive attributes. The first dataset was used for dermatological image classification, where fairness was analysed with respect to Fitzpatrick skin type groups. The second dataset was used for chest X-ray classification, where fairness was analysed with respect to patient age groups. In both cases, the classification task was formulated as a binary problem, and sensitive groups were defined before model training and evaluation.

\subsubsection{DDI}\label{subsubsec:ddi}

The dermatology experiments used the Diverse Dermatology Images (DDI) dataset, a curated clinical image dataset designed to support the evaluation of dermatology AI models across diverse skin tones \cite{Daneshjou2022DDI}. In this work, the task was formulated as binary classification between malignant and benign lesions. After filtering, malignant cases represented approximately 26.1\% of the dataset, while benign cases represented approximately 73.9\%.

The sensitive attribute was defined using Fitzpatrick skin type (FST). To obtain sufficiently populated subgroups for fairness evaluation, the six Fitzpatrick categories were grouped into three broader groups:

\begin{equation}
\mathcal{G}_{\mathrm{DDI}} =
\{
\mathrm{FST\ I\text{--}II},
\mathrm{FST\ III\text{--}IV},
\mathrm{FST\ V\text{--}VI}
\}.
\end{equation}

The percentage class distribution across Fitzpatrick skin type groups is shown in Figure~\ref{fig:class_distribution}. In DDI, the malignant proportion was 23.6\% for FST I--II, 30.7\% for FST III--IV, and 23.2\% for FST V--VI.

\subsubsection{CheXpert}\label{subsubsec:chexpert}

The radiology experiments used CheXpert-derived chest X-ray data, with pleural effusion selected as the binary target condition \cite{Irvin2019CheXpert}. Uncertain labels for pleural effusion were removed, and the task was formulated as binary classification between no pleural effusion and pleural effusion. After filtering, positive cases represented approximately 40.7\% of the dataset.

The sensitive attribute was patient age. Patients were grouped into three clinically interpretable age ranges:

\begin{equation}
\mathcal{G}_{\mathrm{CheXpert}} =
\{
0\text{--}40,
41\text{--}60,
60+
\}.
\end{equation}

\begin{figure}[htbp]
    \centering
    \includegraphics[width=\textwidth]{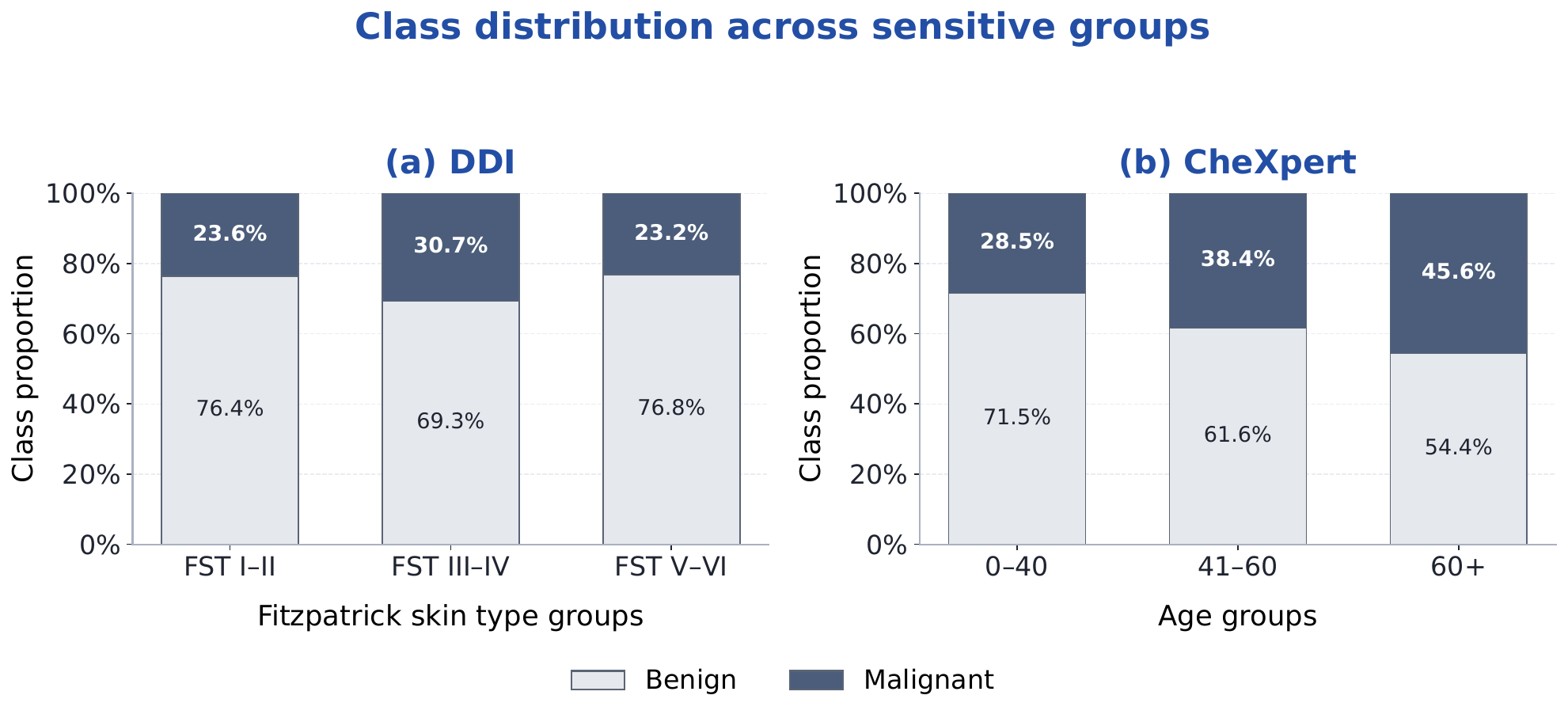}
    \caption{
    \textbf{Class distribution across sensitive groups in the evaluated datasets.}
    Panel (a) shows the percentage distribution of benign and malignant cases across Fitzpatrick skin type groups in DDI. Panel (b) shows the percentage distribution of no pleural effusion and pleural effusion cases across age groups in CheXpert. The figure highlights subgroup-specific differences in class prevalence, supporting the need for fairness-aware evaluation across sensitive groups.
    }
    \label{fig:class_distribution}
\end{figure}

As shown in Figure~\ref{fig:class_distribution}, the proportion of pleural effusion increased with age, from 28.5\% in the 0--40 group to 38.4\% in the 41--60 group and 45.6\% in the 60+ group. This distributional pattern is relevant for fairness analysis because differences in disease prevalence across sensitive groups may affect both predictive performance and fairness metrics. Therefore, subgroup-aware evaluation was used to assess whether model behaviour varied across age groups.

This prevalence pattern also motivates a conceptual distinction that is maintained throughout the analysis. Age-related differences in the probability of pleural effusion are not, in themselves, evidence of algorithmic bias: they reflect genuine epidemiological and physiological structure, since pleural effusion is more frequent in older patients for reasons related to the underlying burden of cardiac, renal, and neoplastic disease~\cite{Petersen2023EqualPerformance,RicciLara2022FairnessImaging}. It is therefore necessary to separate (i) undesirable bias arising from data imbalance, acquisition artefacts, institutional confounding, or spurious correlations, from (ii) legitimate differences arising from disease prevalence and biology~\cite{Tejani2024BiasImagingAI,Xu2024FairMedIAReview}. The two are not addressed by the same remedy. A metric such as demographic parity, which compares positive prediction rates across groups, will register a non-zero gap whenever true prevalence differs across groups, even for a perfectly calibrated and unbiased model~\cite{Dwork2012Fairness,Chouldechova2017FairPrediction}; this is the principal reason why demographic parity is treated here as a supplementary indicator rather than as the main fairness criterion (Section~\ref{subsec:evaluation_metrics}). By the same reasoning, an adversarial intervention such as GRL, which is explicitly designed to remove sensitive-attribute information from the learned representation, may remove not only spurious demographic shortcuts but also epidemiologically informative signal~\cite{Geirhos2020Shortcut,Zhang2018Mitigating}. In medicine, demographic invariance is therefore not synonymous with fairness, and an intervention that enforces it may carry a diagnostic cost that the disparity metric alone does not reveal~\cite{Wachter2021FairnessBias,Petersen2023EqualPerformance}.

\FloatBarrier

\subsection{Feature extraction}\label{subsec:feature_extraction}

Each image was first converted into a fixed-dimensional classical embedding, and this embedding was then used as input to the downstream classical and quantum classifiers. For both DDI and CheXpert, images were processed using the standard ResNet-18 preprocessing pipeline, including resizing, tensor conversion, and normalisation according to the ImageNet convention \cite{Deng2009ImageNet,He2016ResNet}. ResNet-18 was selected as a standard convolutional backbone because residual networks provide robust generic visual representations and are widely used as feature extractors in medical image classification pipelines.

The processed images were passed through a ResNet-18 model pretrained on ImageNet, with the final classification layer removed. The resulting embedding was used as the image representation for the downstream models. This representation strategy ensured that classical and quantum classifiers were trained on a common feature space, reducing the possibility that differences in performance were caused by different image encoders rather than by the classifier family itself.

For the quantum models, the ResNet-18 embeddings were reduced to ten-dimensional feature vectors using principal component analysis (PCA) \cite{Jolliffe2016PCA,Abdi2010PCA}. This step was necessary because both QSVM and QCNN were implemented with ten qubits, and each feature was associated with one qubit. Dimensionality reduction is a common practical requirement in near-term QML workflows, where the number of input features directly affects the number of qubits, circuit width, and simulation cost \cite{Biamonte2017QuantumML,Schuld2021KernelMethods,Franco2026QuantumPhasesSHAP}. In this sense, PCA was used not as a fairness intervention, but as a compact projection that made the ResNet-18 representations compatible with the quantum circuits evaluated in this work.

It should be emphasised that this design places a substantial part of the representational burden---and potentially a substantial part of the subgroup bias---in the classical stage of the pipeline, upstream of any quantum component. All visual information reaching the quantum models has already been filtered by an ImageNet-pretrained ResNet-18 and then compressed to ten principal components. Any bias encoded by the backbone, or any subgroup-relevant signal discarded by the projection, is therefore inherited by the quantum classifiers rather than generated by them~\cite{Larrazabal2020GenderImbalance,Xu2024FairMedIAReview}. The consequences of this design for the attribution of the observed fairness behaviour are discussed in Section~\ref{sec:limitations_future_work}.

\subsection{Classifiers}\label{subsec:classifiers}

\subsubsection{Classical and foundation baselines}\label{subsubsec:classical_baselines}

Two classical classifiers were used as baselines: SVM and MLP. The SVM baseline used a radial basis function kernel and was trained on the extracted image embeddings or their projected representations. The SVM was included as a classical kernel-based model, providing a natural comparison with the QSVM, since both approaches rely on kernel-based decision boundaries \cite{Cortes1995SVM,Scholkopf2002Learning}.

The MLP baseline consisted of a feed-forward neural classifier trained on the extracted embeddings. It provided a nonlinear classical baseline distinct from the SVM. This comparison is relevant because subgroup behaviour may depend not only on the input representation, but also on the classifier architecture, optimisation dynamics, and decision boundary. In addition to these classical baselines, two state-of-the-art foundation models were included as external reference points: PanDerm for the dermatology task and EVA-X for the chest X-ray task \cite{Yan2025PanDerm,Yao2025EVAX}.

Because the classical and quantum classifiers do not necessarily operate on input spaces of identical dimensionality, comparisons between the two families should be read as comparisons between complete pipelines rather than as isolated comparisons between classifier mechanisms~\cite{Schuld2021KernelMethods,SchuldKilloran2019FeatureSpaces}. A difference in performance or in subgroup behaviour between, for example, SVM and QSVM may originate from the dimensionality of the input representation, from the PCA projection, from differences in effective model capacity or regularisation~\cite{Benedetti2019ParameterizedCircuits}, or from the quantum feature map itself, and the present design does not separate these sources. A control in which classical classifiers are trained on exactly the same ten-dimensional PCA representation supplied to the quantum models, at matched capacity, would be required to isolate the contribution of the quantum component; this is identified as a priority for future work in Section~\ref{sec:limitations_future_work}.

\subsubsection{Quantum support vector machine}\label{subsubsec:qsvm}

The QSVM was implemented as a quantum-kernel classifier \cite{Havlicek2019Supervised,Schuld2021KernelMethods}. Each ResNet-18 embedding was first projected into a ten-dimensional feature vector using PCA,

\begin{equation}
x =
(x_1, x_2, \ldots, x_{10})
\in \mathbb{R}^{10},
\end{equation}

where each feature was associated with one qubit. The quantum feature map used ten qubits and was simulated using PennyLane \cite{Bergholm2018PennyLane}.

The input vector was encoded using an angle-embedding strategy based on single-qubit rotations. First, the features were encoded through rotations around the $Y$ axis. Then, additional feature-dependent rotations and entangling operations were applied. The feature map can be written schematically as

\begin{equation}
|\phi(x)\rangle
=
U_{\mathrm{FM}}(x)
|0\rangle^{\otimes 10},
\end{equation}

where $U_{\mathrm{FM}}(x)$ denotes the feature-map unitary. In this work, $U_{\mathrm{FM}}(x)$ combined angle embedding with a ring of CNOT gates and feature-dependent rotations of the form $R_Z(x_i)$ and $R_X(0.5x_i)$. The feature map can be represented as

\begin{equation}
U_{\mathrm{FM}}(x)
=
\left[
\prod_{i=1}^{10}
R_X(0.5x_i) R_Z(x_i)
\right]
U_{\mathrm{ring}}
\left[
\prod_{i=1}^{10}
R_Y(x_i)
\right],
\end{equation}

where $U_{\mathrm{ring}}$ denotes the CNOT ring entangling layer. The CNOT ring connects neighbouring qubits and closes the chain by entangling the last qubit with the first one. This structure was used to introduce feature-dependent rotations and local entanglement while keeping the circuit shallow enough for statevector simulation.

Given two samples $x_i$ and $x_j$, the quantum kernel was computed using the fidelity between the corresponding quantum states:

\begin{equation}
K(x_i,x_j)
=
\left|
\langle \phi(x_i) | \phi(x_j) \rangle
\right|^2.
\label{eq:method_quantum_kernel}
\end{equation}

The resulting kernel matrix was then used in an SVM classifier. The main computational cost of the QSVM is associated with the construction of the kernel matrix. For a training set with $N$ samples, the number of pairwise kernel evaluations scales as $\mathcal{O}(N^2)$, while evaluating new samples requires additional comparisons with the training set. This quadratic scaling is a central computational difference between quantum-kernel methods and trainable variational classifiers, and it motivates the use of compact input representations in practical QML studies \cite{Schuld2021KernelMethods,Franco2026QuantumPhasesSHAP}.

\subsubsection{Quantum convolutional neural network}\label{subsubsec:qcnn}

The QCNN was implemented as a trainable variational quantum classifier using ten qubits and simulated with PennyLane \cite{Bergholm2018PennyLane}. As in the QSVM, each input image embedding was projected into a ten-dimensional feature vector using PCA. The input was encoded using angle embedding with rotations around the $Y$ axis:

\begin{equation}
U_{\mathrm{enc}}(x)
=
\prod_{i=1}^{10}
R_Y(x_i).
\end{equation}

After encoding, the quantum state was processed by a trainable variational circuit inspired by the QCNN architecture \cite{Cong2019QCNN}. QCNNs were originally proposed as quantum analogues of convolutional architectures, using local trainable operations and pooling-like mechanisms to extract information from quantum states. In this work, the ansatz consisted of trainable quantum layers arranged in convolution-like and pooling-like blocks, combining parametrised single-qubit rotations with entangling operations. The purpose of this ansatz was to learn a nonlinear decision function from the encoded ten-dimensional representation.

A generic prediction from the QCNN can be written as

\begin{equation}
f_{\theta}(x)
=
\langle 0|^{\otimes 10}
U_{\mathrm{enc}}^{\dagger}(x)
U_{\mathrm{QCNN}}^{\dagger}(\theta)
M
U_{\mathrm{QCNN}}(\theta)
U_{\mathrm{enc}}(x)
|0\rangle^{\otimes 10},
\label{eq:method_qcnn_prediction}
\end{equation}

where $U_{\mathrm{enc}}(x)$ is the angle-embedding unitary, $U_{\mathrm{QCNN}}(\theta)$ is the trainable variational circuit, $M$ is the measurement observable, and $\theta$ denotes the trainable parameters.

Unlike the QSVM, the QCNN does not require the construction of a full pairwise kernel matrix. Instead, its computational cost is dominated by repeated circuit evaluations during training. The total cost therefore depends on the number of samples, the number of trainable parameters, the number of optimisation steps, and the number of seeds. This makes the QCNN computationally different from the QSVM: the QSVM concentrates cost in kernel-matrix construction, while the QCNN concentrates cost in variational optimisation \cite{Cerezo2021VQA,Tilly2022VQAReview}.

\subsection{Simulation protocol and reproducibility}\label{subsec:simulation_protocol}

All quantum models were simulated using PennyLane \cite{Bergholm2018PennyLane}. Both QSVM and QCNN used ten qubits, but they differed in the role of the quantum circuit. In the QSVM, the circuit was used as a fidelity-kernel feature map, and the final classifier was a kernel-based SVM. In the QCNN, the circuit was a trainable variational model optimised directly for the classification task.

All quantum computations were performed in classical statevector simulation. No experiment was executed on quantum hardware, and no claim of quantum advantage---in predictive performance, in fairness behaviour, or in computational efficiency---is made or tested in this work~\cite{Preskill2018NISQ,Cerezo2021VQA}. The quantum models are studied as one component of a hybrid classical--quantum pipeline, and the object of the audit is the behaviour of that pipeline as a whole.

For trainable neural and variational components, including the MLP, QCNN, and adversarial branches used in GRL, optimisation was performed with Adam \cite{Kingma2015Adam}. The learning rate was set to $10^{-3}$, weight decay to $10^{-4}$, the maximum number of training epochs to 25, and early stopping patience to 5 epochs.

All experiments were repeated using the five seeds $\{13,42,123,777,2024\}$. For each seed, the dataset was split into training, validation, and test sets using a 60/20/20 split, and the models were trained and evaluated independently. The reported numerical values correspond to the held-out test-set results. The tables report point estimates averaged across seeds, with the corresponding standard deviation computed across the five runs. Therefore, the uncertainty values reflect variability across independent experimental repetitions rather than resampling-based confidence intervals.

\subsection{Fairness mitigation strategies}\label{subsec:fairness_mitigation}

\begin{figure}[htbp]
    \centering
    \includegraphics[width=\textwidth]{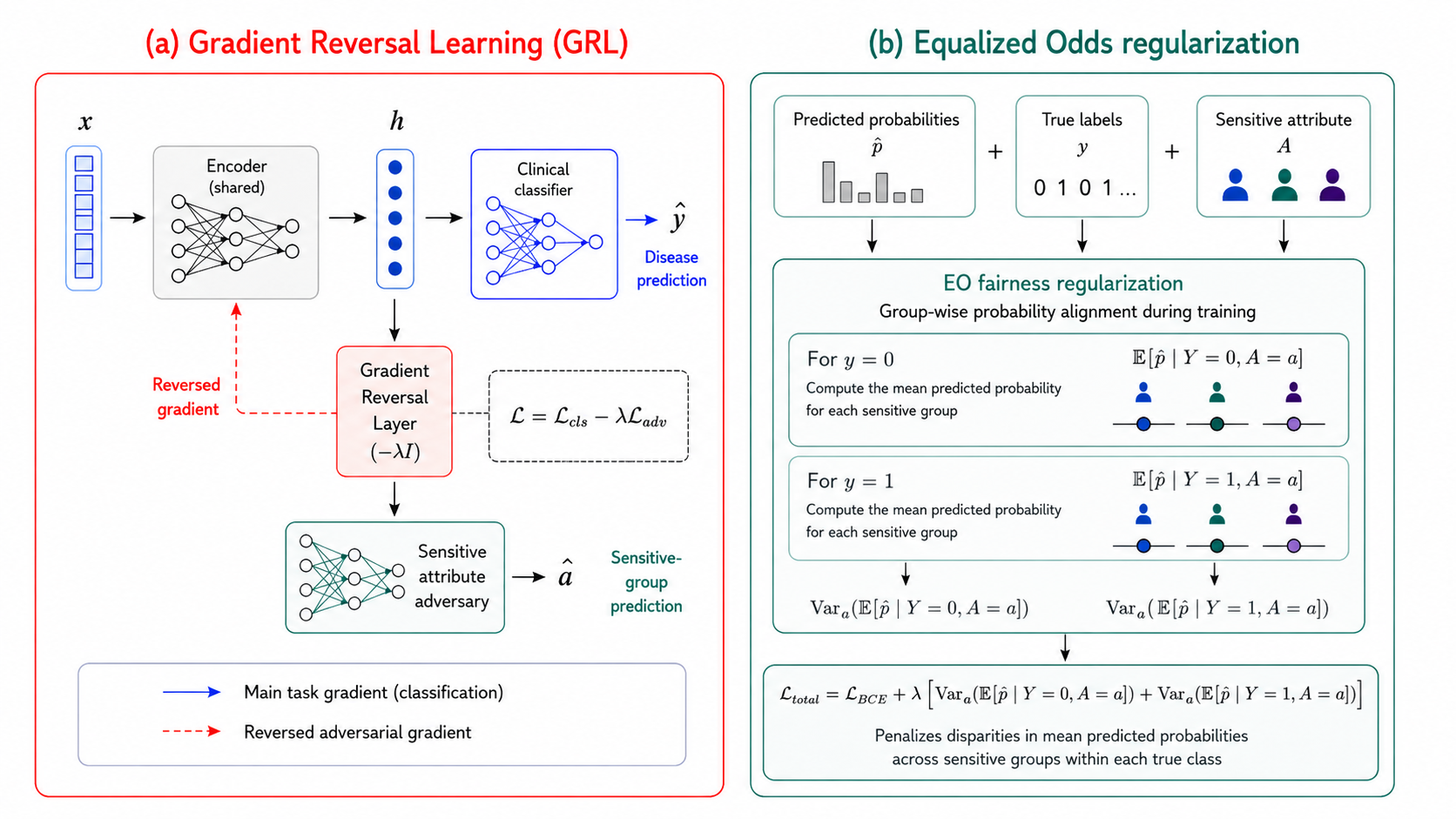}
    \caption{
    \textbf{Fairness mitigation strategies used in the proposed evaluation.}
    Panel (a) shows Gradient Reversal Learning (GRL), in which a shared encoder produces a representation used by both the clinical classifier and a sensitive-attribute adversary. The gradient reversal layer gradually reverses the adversarial gradient according to a scheduled coefficient, encouraging the representation to preserve disease-relevant information while reducing information about the sensitive group. Panel (b) shows the differentiable Equalized Odds (EO) penalty used during training. The predicted probabilities are separated by true class and sensitive group, group-wise mean probabilities are computed within each class, and the variance of these group-wise means is added to the binary cross-entropy loss. This encourages the model to reduce group-dependent differences in prediction behaviour for both negative and positive cases.
    }
    \label{fig:method_bias}
\end{figure}

Four evaluation settings were considered: baseline, Equalized Odds (EO), Gradient Reversal Learning (GRL), and the combined GRL+EO strategy. In this work, EO was implemented directly during training as a differentiable fairness regularisation term added to the classification loss. Figure~\ref{fig:method_bias} summarises the two mitigation mechanisms used in this work. Panel (a) illustrates GRL as an adversarial representation-learning strategy, while Panel (b) illustrates the EO differentiable penalty used to reduce group-dependent differences in predicted probabilities within each true class.

The baseline setting trained each classifier without an explicit fairness intervention. This setting was used as the reference point for analysing predictive performance and subgroup disparity.

EO was implemented as a differentiable regularisation term during training. Equalized Odds aims to make the model prediction conditionally independent of the sensitive attribute given the true class label \cite{Hardt2016Equality}. In the differentiable formulation used here, the model first computes the predicted probability of the positive class for each sample. These probabilities are then separated by the true label, $Y=0$ and $Y=1$, and, within each class, by the sensitive group $A=a$. For each class, the mean predicted probability is computed for each sensitive group. The EO penalty is defined as the variance of these group-wise means, summed across the negative and positive classes.

Let $\hat{p}_\theta(x)$ denote the predicted probability of the positive class. The EO-regularised loss was defined as

\begin{equation}
\mathcal{L}_{\mathrm{EO}}
=
\mathcal{L}_{\mathrm{BCE}}
+
\lambda_{\mathrm{EO}}
\mathcal{R}_{\mathrm{EO}},
\label{eq:eo_differentiable_loss}
\end{equation}

where

\begin{equation}
\begin{aligned}
\mathcal{R}_{\mathrm{EO}}
=&\;
\operatorname{Var}_{a}
\left(
\mathbb{E}\left[\hat{p}_\theta(x) \mid Y=0, A=a\right]
\right)
\\
&+
\operatorname{Var}_{a}
\left(
\mathbb{E}\left[\hat{p}_\theta(x) \mid Y=1, A=a\right]
\right).
\end{aligned}
\label{eq:eo_penalty}
\end{equation}

Here, $\mathcal{L}_{\mathrm{BCE}}$ is the binary cross-entropy loss, $A$ is the sensitive attribute, and $a$ indexes the sensitive groups. In all EO experiments, the regularisation weight was fixed at $\lambda_{\mathrm{EO}}=0.50$. This formulation penalises differences in the average predicted probability across sensitive groups separately for benign and malignant cases in DDI, and separately for no pleural effusion and pleural effusion cases in CheXpert. Therefore, the EO term directly affects the optimisation objective and encourages the learned classifier to reduce class-conditional group disparities during training.

GRL was applied as an adversarial representation-learning strategy \cite{Ganin2016DomainAdversarial}. The use of GRL is also consistent with recent fairness-aware medical imaging pipelines, where adversarial branches have been used to discourage learned representations from retaining demographic information that may act as a shortcut for the main diagnostic task \cite{Parikh2026FairLungGRL}. In this setting, the shared representation is used simultaneously by a clinical classifier, which predicts the disease label, and by a sensitive-attribute adversary, which attempts to predict the sensitive group. The adversary was implemented as a small multilayer perceptron applied after the gradient reversal layer:

\begin{equation}
h
\rightarrow
\mathrm{GRL}_{\lambda_{\mathrm{GRL}}(t)}
\rightarrow
\mathrm{Linear}(d,64)
\rightarrow
\mathrm{ReLU}
\rightarrow
\mathrm{Linear}(64,3)
\rightarrow
z_A,
\end{equation}

where $h$ is the learned representation, $d$ is its input dimension, and $z_A$ denotes the logits for the three sensitive groups. The same three-output adversarial structure was used for both DDI, where the groups correspond to FST I--II, FST III--IV, and FST V--VI, and CheXpert, where the groups correspond to 0--40, 41--60, and 60+.

It is worth making explicit what this objective does and does not assume. The adversarial term treats sensitive-attribute information in the representation as undesirable by construction, without distinguishing whether that information acts as a spurious shortcut or as legitimate clinical signal~\cite{Zhang2018Mitigating,Geirhos2020Shortcut}. In CheXpert, where the prevalence of pleural effusion varies markedly with age (Section~\ref{subsubsec:chexpert}), age is not only a demographic descriptor but also a genuine epidemiological predictor of the target condition~\cite{Petersen2023EqualPerformance}. Enforcing invariance to it may therefore suppress diagnostic signal alongside any shortcut. This provides a mechanistic expectation that GRL-based settings may trade global discrimination for representational invariance~\cite{Zietlow2022LevelingDown,Zhang2018Mitigating}, and it is one reason why the mitigated configurations are evaluated below against an explicit non-inferiority margin rather than by disparity reduction alone.

The GRL coefficient was dynamic. During the first five epochs, a warm-up stage was used and the gradient reversal coefficient was fixed at zero. From epoch 6 to epoch 25, the coefficient increased linearly from 0 to a maximum value of 1.00:

\begin{equation}
\lambda_{\mathrm{GRL}}(t)
=
\begin{cases}
0, & t \leq 5,\\
\dfrac{t-6}{25-6}, & 6 \leq t \leq 25,
\end{cases}
\label{eq:grl_schedule}
\end{equation}

where $t$ denotes the training epoch. This schedule was used to avoid forcing sensitive-attribute invariance before the clinical classifier had learned a useful initial representation. The adversarial loss had a separate weight $\alpha_{\mathrm{adv}}=0.15$. From the perspective of the encoder, the effective GRL objective can be written as

\begin{equation}
\mathcal{L}_{\mathrm{GRL}}^{\mathrm{enc}}(t)
=
\mathcal{L}_{\mathrm{cls}}
-
\alpha_{\mathrm{adv}}
\lambda_{\mathrm{GRL}}(t)
\mathcal{L}_{\mathrm{adv}},
\label{eq:grl_loss}
\end{equation}

where $\mathcal{L}_{\mathrm{cls}}$ is the clinical classification loss and $\mathcal{L}_{\mathrm{adv}}$ is the adversarial cross-entropy loss for sensitive-group prediction. Thus, the GRL gradually reverses the adversarial gradient during training, encouraging the learned representation to contain less information about the sensitive group while retaining information useful for the clinical task.

The combined GRL+EO strategy integrated both mechanisms during training. In this setting, the representation was trained with the scheduled adversarial GRL objective, while the classifier was also regularised by the differentiable EO penalty. Conceptually, GRL acts at the representation level by discouraging sensitive-group information in the learned features, whereas EO acts at the prediction level by penalising class-conditional differences in mean predicted probabilities across sensitive groups. The combined GRL+EO objective was defined as

\begin{equation}
\begin{aligned}
\mathcal{L}_{\mathrm{GRL+EO}}(t)
=&\;
\mathcal{L}_{\mathrm{cls}}
-
\alpha_{\mathrm{adv}}
\lambda_{\mathrm{GRL}}(t)
\mathcal{L}_{\mathrm{adv}}
+
\lambda_{\mathrm{EO}}
\mathcal{R}_{\mathrm{EO}},
\end{aligned}
\label{eq:grl_eo_loss}
\end{equation}

with $\alpha_{\mathrm{adv}}=0.15$, $\lambda_{\mathrm{EO}}=0.50$, and $\lambda_{\mathrm{GRL}}(t)$ defined by Eq.~\ref{eq:grl_schedule}. These interventions were not assumed to improve all fairness metrics simultaneously. Instead, they were used to analyse performance--fairness trade-offs and to test how differentiable mitigation strategies commonly used or inspired by classical machine learning behave when adapted to QML workflows.

For QSVM-based mitigation settings, the fairness-aware objectives were applied before quantum-kernel construction. Since the QSVM relies on a fixed kernel matrix rather than an end-to-end differentiable classifier, EO, GRL, and GRL+EO were implemented at the representation stage. A lightweight differentiable projection module was trained to map the ResNet-18 embeddings into the ten-dimensional quantum input space using the corresponding fairness-aware objective. The resulting ten-dimensional representations were then fixed and used to compute the QSVM fidelity kernel. Thus, the mitigation strategies did not alter the SVM optimisation directly, but changed the feature representation on which the quantum kernel was evaluated.

\subsection{Evaluation metrics}\label{subsec:evaluation_metrics}

The models were evaluated using predictive performance metrics and subgroup-aware fairness metrics. Predictive performance was measured using AUC, balanced accuracy, and F1-score. These metrics were selected because they capture complementary aspects of binary classification performance, including ranking quality, class-balanced performance, and the balance between precision and recall \cite{Fawcett2006ROC,Saito2015PrecisionRecall}.

Balanced accuracy was computed as the average between sensitivity and specificity:

\begin{equation}
\mathrm{Balanced\ Accuracy}
=
\frac{1}{2}
\left(
\mathrm{TPR}
+
\mathrm{TNR}
\right),
\label{eq:balanced_accuracy}
\end{equation}

where $\mathrm{TPR}$ is the true positive rate and $\mathrm{TNR}$ is the true negative rate.

Fairness was evaluated by comparing model behaviour across sensitive groups. Let $\mathcal{G}$ denote the set of sensitive groups. The main subgroup-disparity metric used in this work was the group-wise AUC gap:

\begin{equation}
\Delta_{\mathrm{AUC}}
=
\max_{g \in \mathcal{G}}
\mathrm{AUC}_{g}
-
\min_{g \in \mathcal{G}}
\mathrm{AUC}_{g}.
\label{eq:method_auc_gap}
\end{equation}

The AUC gap was used as the main fairness-oriented metric because it measures differences in discriminative performance across groups without depending on a single fixed decision threshold. This is particularly relevant in medical classification tasks, where disease prevalence may differ across sensitive groups.

The AUC gap is, however, a difference between two extreme subgroup values, and its magnitude alone does not indicate how that difference was obtained. To make this explicit, two additional quantities are reported for every mitigated configuration $m$ relative to the corresponding unmitigated baseline $b$ of the same model family. The first is the change in global discrimination,

\begin{equation}
\delta_{\mathrm{AUC}}
=
\mathrm{AUC}^{(m)}
-
\mathrm{AUC}^{(b)},
\label{eq:delta_auc_global}
\end{equation}

and the second is the change in subgroup disparity,

\begin{equation}
\delta_{\Delta}
=
\Delta_{\mathrm{AUC}}^{(m)}
-
\Delta_{\mathrm{AUC}}^{(b)}.
\label{eq:delta_gap}
\end{equation}

A mitigation strategy is regarded as favourable only when it satisfies both of the following conditions simultaneously:

\begin{equation}
\delta_{\Delta} < 0
\qquad \text{and} \qquad
\delta_{\mathrm{AUC}} \geq -\,\varepsilon,
\label{eq:non_inferiority}
\end{equation}

where $\varepsilon$ is a pre-specified non-inferiority margin on global discrimination. Throughout this work $\varepsilon = 0.02$ AUC points. This value was fixed before the analysis and was chosen to be of the same order as, but not smaller than, the seed-to-seed variability reported in Tables~\ref{tab:performance_chexpert} and~\ref{tab:performance_ddi}, so that a configuration is not declared acceptable on the basis of a difference indistinguishable from run-to-run noise~\cite{Piaggio2012CONSORT}. The margin is a methodological convention adopted for this audit and not a validated clinical threshold; in a deployment setting it would have to be established for the specific task, population, and care pathway, in consultation with domain experts~\cite{Piaggio2012CONSORT,Ueda2024FairnessHealthcare}.

The rationale for Eq.~\ref{eq:non_inferiority} is that the first condition alone is degenerate. A model whose discrimination collapses towards chance in every subgroup would satisfy $\delta_{\Delta} < 0$ and would be assigned a near-zero disparity, while being clinically useless~\cite{Mittelstadt2024LevellingDown,Zietlow2022LevelingDown}. Requiring joint satisfaction of both conditions excludes this degenerate solution and distinguishes disparity reduction achieved by raising the worst-performing group from disparity reduction achieved by lowering the best-performing one. It should be noted that Eq.~\ref{eq:non_inferiority} constrains \emph{global} discrimination, because subgroup-level AUC values were computed for the baseline configurations only; the stronger per-subgroup form of the criterion, in which no individual group may deteriorate beyond $\varepsilon$, is identified as future work in Section~\ref{sec:limitations_future_work}.

As a complementary metric, the demographic parity gap was computed as

\begin{equation}
\Delta_{\mathrm{DP}}
=
\max_{g \in \mathcal{G}}
P(\hat{Y}=1 \mid A=g)
-
\min_{g \in \mathcal{G}}
P(\hat{Y}=1 \mid A=g),
\label{eq:method_dp_gap}
\end{equation}

where $\hat{Y}$ is the predicted label and $A$ is the sensitive attribute. Since demographic parity can be affected by true prevalence differences across clinical groups, it was interpreted as a supplementary indicator rather than as the main fairness criterion \cite{Hardt2016Equality,Barocas2023FairnessML,Mehrabi2021Survey}.

These metrics were analysed jointly because predictive performance and subgroup disparity may lead to different conclusions, and because different fairness criteria may be statistically incompatible in realistic settings \cite{Chouldechova2017FairPrediction,Kleinberg2017InherentTradeoffs}. Therefore, the evaluation focused on the interaction between predictive performance, AUC gap, and demographic parity gap, rather than selecting a single metric as definitive.

Finally, all metrics reported here are threshold-independent (AUC, AUC gap) or computed at a single calibrated operating point (balanced accuracy, F1-score, demographic parity gap). They therefore characterise ranking quality and aggregate error behaviour, but they do not weight the two error types according to their clinical consequences~\cite{Fawcett2006ROC,Saito2015PrecisionRecall}. In the tasks considered here, false negatives and false positives are not clinically interchangeable: a missed malignant lesion in DDI and a missed pleural effusion in CheXpert carry consequences of a different nature and magnitude from those of a false alarm. A mitigation strategy that reduces the AUC gap while shifting the error profile towards false negatives could therefore be undesirable in practice despite appearing more equitable under the metrics reported here~\cite{Petersen2023EqualPerformance}. A complete clinical characterisation would require subgroup-level sensitivity, specificity, positive predictive value, and negative predictive value at clinically motivated operating points~\cite{AltmanBland1994PredictiveValues}, together with a net-benefit analysis such as decision curve analysis~\cite{Vickers2006DCA,Vickers2016NetBenefit}; this is outside the scope of the present audit and is stated as a limitation in Section~\ref{sec:limitations_future_work}.

\section{Results}\label{sec:results}

This section presents the empirical results of the proposed fairness audit. The analysis is organised into  six complementary parts. First, global predictive performance is reported for CheXpert and DDI. Second, subgroup-specific ROC curves are used to inspect how baseline classifiers behave across sensitive groups. Third, performance--fairness trade-offs are analysed using AUC gap as the main subgroup-disparity metric. Fourth, an AUC gap heatmap provides a direct comparison of threshold-independent disparity across models and mitigation strategies.  Fifth, demographic parity gap is reported as a complementary fairness indicator to analyse differences in positive prediction rates across sensitive groups. Finally, every mitigated configuration is evaluated against the non-inferiority criterion of Eq.~\ref{eq:non_inferiority}, in order to separate disparity reductions that preserve global discrimination from those that do not.

\subsection{Global predictive performance}\label{subsec:global_performance}

Table~\ref{tab:performance_chexpert} reports the global predictive performance on CheXpert. The EVA-X foundation model is included as a reference baseline and presented the highest aggregate point estimates, with AUC, balanced accuracy, and F1-score values of 0.925, 0.846, and 0.782, respectively. Among the downstream classifiers, the quantum models appeared to be competitive with the classical baselines in this experimental setting. In particular, QCNN with GRL presented the highest non-foundation point estimates, with an AUC of $0.820 \pm 0.029$, balanced accuracy of $0.771 \pm 0.030$, and F1-score of $0.662 \pm 0.044$.

\begin{table}[htbp]
\centering
\scriptsize
\setlength{\tabcolsep}{5pt}
\renewcommand{\arraystretch}{1.15}
\caption{
Global predictive performance on CheXpert across models and fairness mitigation strategies.
Values are reported as point estimates with standard deviations across seeds when available.
The foundation model is reported as a reference baseline and marked with ``---'' because no fairness mitigation strategy was applied.
Bold values indicate the highest point estimate among non-foundation models for each metric.
Bold marks the highest point estimate only; because the reported deviations are seed-to-seed variability rather than confidence intervals, and are not paired across models, differences between configurations with overlapping deviations should not be read as evidence of superiority~\cite{AltmanBland1995Absence}.
}
\label{tab:performance_chexpert}
\begin{tabular}{ccccc}
\toprule
\textbf{Model} & \textbf{Mitigation} & \textbf{AUC} & \textbf{Balanced acc.} & \textbf{F1-score} \\
\midrule
\textbf{EVA-X (foundation)} & \textbf{---} & $0.925 \pm 0.029$ & $0.846 \pm 0.032$ & $0.782 \pm 0.043$ \\
\midrule
SVM & Baseline & $0.783 \pm 0.032$ & $0.732 \pm 0.031$ & $0.611 \pm 0.044$ \\
SVM & Eq. Odds & $0.797 \pm 0.030$ & $0.735 \pm 0.031$ & $0.617 \pm 0.045$ \\
SVM & GRL & $0.783 \pm 0.031$ & $0.737 \pm 0.030$ & $0.618 \pm 0.044$ \\
SVM & GRL+EO & $0.789 \pm 0.032$ & $0.709 \pm 0.031$ & $0.582 \pm 0.044$ \\
MLP & Baseline & $0.805 \pm 0.029$ & $0.753 \pm 0.031$ & $0.640 \pm 0.046$ \\
MLP & Eq. Odds & $0.797 \pm 0.029$ & $0.746 \pm 0.030$ & $0.631 \pm 0.045$ \\
MLP & GRL & $0.784 \pm 0.031$ & $0.735 \pm 0.030$ & $0.615 \pm 0.043$ \\
MLP & GRL+EO & $0.788 \pm 0.031$ & $0.725 \pm 0.031$ & $0.603 \pm 0.045$ \\
QSVM & Baseline & $0.784 \pm 0.031$ & $0.721 \pm 0.031$ & $0.599 \pm 0.046$ \\
QSVM & Eq. Odds & $0.817 \pm 0.030$ & $0.756 \pm 0.030$ & $0.641 \pm 0.043$ \\
QSVM & GRL & $0.779 \pm 0.031$ & $0.723 \pm 0.031$ & $0.600 \pm 0.045$ \\
QSVM & GRL+EO & $0.800 \pm 0.031$ & $0.729 \pm 0.031$ & $0.606 \pm 0.043$ \\
QCNN & Baseline & $0.790 \pm 0.031$ & $0.691 \pm 0.032$ & $0.561 \pm 0.045$ \\
QCNN & Eq. Odds & $0.786 \pm 0.032$ & $0.729 \pm 0.031$ & $0.608 \pm 0.044$ \\
QCNN & GRL & $\mathbf{0.820 \pm 0.029}$ & $\mathbf{0.771 \pm 0.030}$ & $\mathbf{0.662 \pm 0.044}$ \\
QCNN & GRL+EO & $0.806 \pm 0.030$ & $0.746 \pm 0.029$ & $0.626 \pm 0.043$ \\
\bottomrule
\end{tabular}

\vspace{0.4em}
\begin{minipage}{0.96\textwidth}
\footnotesize
\textit{Note:} Eq. Odds denotes differentiable Equalized Odds regularisation; GRL denotes Gradient Reversal Learning; GRL+EO denotes the combination of GRL and Equalized Odds. The foundation model is reported as a reference baseline and is marked with ``---'' because no fairness mitigation strategy was applied to it. Bold values indicate the highest point estimate among non-foundation models.
\end{minipage}
\end{table}

Table~\ref{tab:performance_ddi} reports the corresponding results for DDI. PanDerm is included as the foundation-model reference and presented an AUC of 0.814, balanced accuracy of 0.766, and F1-score of 0.640. Among the downstream models, QSVM with GRL+EO presented the highest overall non-foundation point estimates, with an AUC of $0.808 \pm 0.051$, balanced accuracy of $0.770 \pm 0.046$, and F1-score of $0.651 \pm 0.070$.

These point estimates are close to those of the PanDerm reference across all three metrics (AUC $0.808$ vs.\ $0.814$; balanced accuracy $0.770$ vs.\ $0.766$; F1-score $0.651$ vs.\ $0.640$). The differences are, however, an order of magnitude smaller than the corresponding seed-to-seed deviations, which are of the order of $0.05$ for AUC and $0.06$--$0.07$ for F1-score, and the two models were not compared under a paired statistical test~\cite{DeLong1988}. The appropriate reading is therefore that QSVM with GRL+EO attains performance comparable to the foundation-model reference on this dataset, and not that it outperforms it on any individual metric. Establishing superiority, inferiority, or formal equivalence would require paired testing and resampling-based confidence intervals~\cite{DeLong1988,Efron1993Bootstrap}, which were not computed in this study (Section~\ref{sec:limitations_future_work}). What can be stated is the weaker but still substantive claim that a fairness-aware quantum classifier built on ten principal components remains in the same performance range as a domain-specific dermatology foundation model in this setting.

\begin{table}[htbp]
\centering
\scriptsize
\setlength{\tabcolsep}{4pt}
\renewcommand{\arraystretch}{1.15}
\caption{
Global predictive performance on DDI across models and fairness mitigation strategies.
Values are reported as point estimates with standard deviations across seeds when available.
The foundation model is reported as a reference baseline and marked with ``---'' because no fairness mitigation strategy was applied.
Bold values indicate the highest point estimate among non-foundation models for each metric.
As in Table~\ref{tab:performance_chexpert}, overlapping deviations do not support claims of superiority between configurations.
}
\label{tab:performance_ddi}
\begin{tabular}{ccccc}
\toprule
\textbf{Model} & \textbf{Mitigation} & \textbf{AUC} & \textbf{Balanced acc.} & \textbf{F1-score} \\
\midrule
\textbf{PanDerm (foundation)} & \textbf{---} & $0.814 \pm 0.051$ & $0.766 \pm 0.043$ & $0.640 \pm 0.064$ \\
\midrule
SVM & Baseline & $0.771 \pm 0.049$ & $0.724 \pm 0.042$ & $0.568 \pm 0.061$ \\
SVM & Eq. Odds & $0.761 \pm 0.050$ & $0.704 \pm 0.047$ & $0.549 \pm 0.065$ \\
SVM & GRL & $0.755 \pm 0.048$ & $0.687 \pm 0.045$ & $0.533 \pm 0.068$ \\
SVM & GRL+EO & $0.733 \pm 0.054$ & $0.654 \pm 0.045$ & $0.484 \pm 0.078$ \\
MLP & Baseline & $0.776 \pm 0.047$ & $0.693 \pm 0.046$ & $0.543 \pm 0.072$ \\
MLP & Eq. Odds & $0.769 \pm 0.052$ & $0.761 \pm 0.044$ & $0.632 \pm 0.066$ \\
MLP & GRL & $0.781 \pm 0.049$ & $0.730 \pm 0.045$ & $0.585 \pm 0.065$ \\
MLP & GRL+EO & $0.775 \pm 0.048$ & $0.710 \pm 0.046$ & $0.558 \pm 0.065$ \\
QSVM & Baseline & $0.793 \pm 0.045$ & $0.766 \pm 0.044$ & $0.640 \pm 0.067$ \\
QSVM & Eq. Odds & $0.726 \pm 0.055$ & $0.702 \pm 0.047$ & $0.553 \pm 0.070$ \\
QSVM & GRL & $0.802 \pm 0.043$ & $0.705 \pm 0.045$ & $0.552 \pm 0.064$ \\
QSVM & GRL+EO & $\mathbf{0.808 \pm 0.051}$ & $\mathbf{0.770 \pm 0.046}$ & $\mathbf{0.651 \pm 0.070}$ \\
QCNN & Baseline & $0.737 \pm 0.048$ & $0.682 \pm 0.049$ & $0.528 \pm 0.074$ \\
QCNN & Eq. Odds & $0.669 \pm 0.055$ & $0.658 \pm 0.044$ & $0.500 \pm 0.062$ \\
QCNN & GRL & $0.695 \pm 0.052$ & $0.652 \pm 0.046$ & $0.487 \pm 0.066$ \\
QCNN & GRL+EO & $0.780 \pm 0.043$ & $0.731 \pm 0.046$ & $0.590 \pm 0.069$ \\
\bottomrule
\end{tabular}

\vspace{0.4em}
\begin{minipage}{0.96\textwidth}
\footnotesize
\textit{Note:} Eq. Odds denotes differentiable Equalized Odds regularisation; GRL denotes Gradient Reversal Learning; GRL+EO denotes the combination of GRL and Equalized Odds. The foundation model is reported as a reference baseline and is marked with ``---'' because no fairness mitigation strategy was applied to it. Bold values indicate the highest point estimate among non-foundation models.
\end{minipage}
\end{table}

Overall, Tables~\ref{tab:performance_chexpert} and~\ref{tab:performance_ddi} suggest that QML classifiers may be competitive with classical baselines in the evaluated pipelines. The foundation models remain strong reference points, especially on CheXpert, and global metrics alone do not determine whether a model behaves consistently across sensitive groups. For this reason, subgroup-level ROC curves and fairness gaps were analysed next.

\FloatBarrier

\subsection{Subgroup ROC analysis}\label{subsec:roc_results}

\begin{figure}[htbp]
    \centering
    \includegraphics[width=\textwidth]{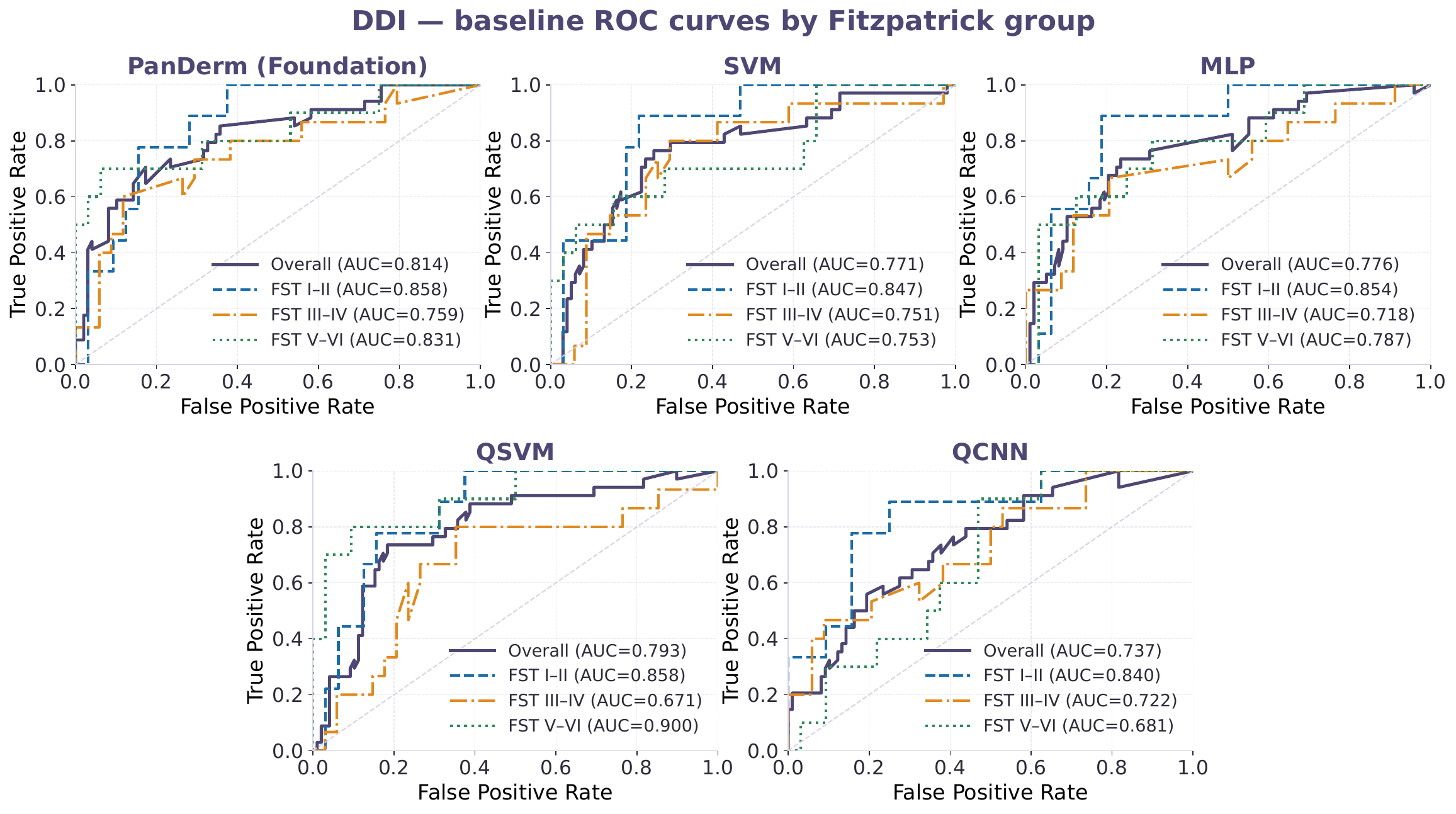}
    \caption{
    \textbf{Baseline ROC curves on DDI by Fitzpatrick skin type group.}
    Each panel shows the overall ROC curve and subgroup-specific curves for FST I--II, FST III--IV, and FST V--VI. The figure illustrates that global predictive performance may conceal subgroup-level variation in model discrimination.
    }
    \label{fig:roc_ddi}
\end{figure}

Figure~\ref{fig:roc_ddi} shows the baseline ROC curves for DDI, including the foundation model, classical baselines, and quantum classifiers. Each panel compares the overall ROC curve with subgroup-specific curves for Fitzpatrick skin type groups. The results suggest that global AUC values may hide relevant subgroup differences. For the PanDerm foundation model, for example, the AUC was 0.858 for FST I--II, 0.759 for FST III--IV, and 0.811 for FST V--VI, corresponding to an AUC gap of 0.099 between the best- and worst-performing groups. Thus, even this strong dermatology reference model appears to exhibit measurable variation across skin type groups.

A similar pattern may be observed for the downstream classifiers. For the classical models, SVM obtained subgroup AUCs of 0.847, 0.751, and 0.753 for FST I--II, FST III--IV, and FST V--VI, respectively, while MLP obtained 0.854, 0.718, and 0.787. These values suggest that the intermediate Fitzpatrick group, FST III--IV, may be the weakest subgroup for the classical baselines in this setting. The quantum models also exhibited subgroup variation. QSVM obtained AUCs of 0.858 for FST I--II, 0.671 for FST III--IV, and 0.900 for FST V--VI, corresponding to the largest baseline subgroup difference among the DDI models. QCNN obtained AUCs of 0.840, 0.722, and 0.681, respectively, suggesting lower discrimination for FST V--VI. These results indicate that the quantum classifiers should not be assumed to behave uniformly across skin type groups under the baseline setting.

\begin{figure}[htbp]
    \centering
    \includegraphics[width=\textwidth]{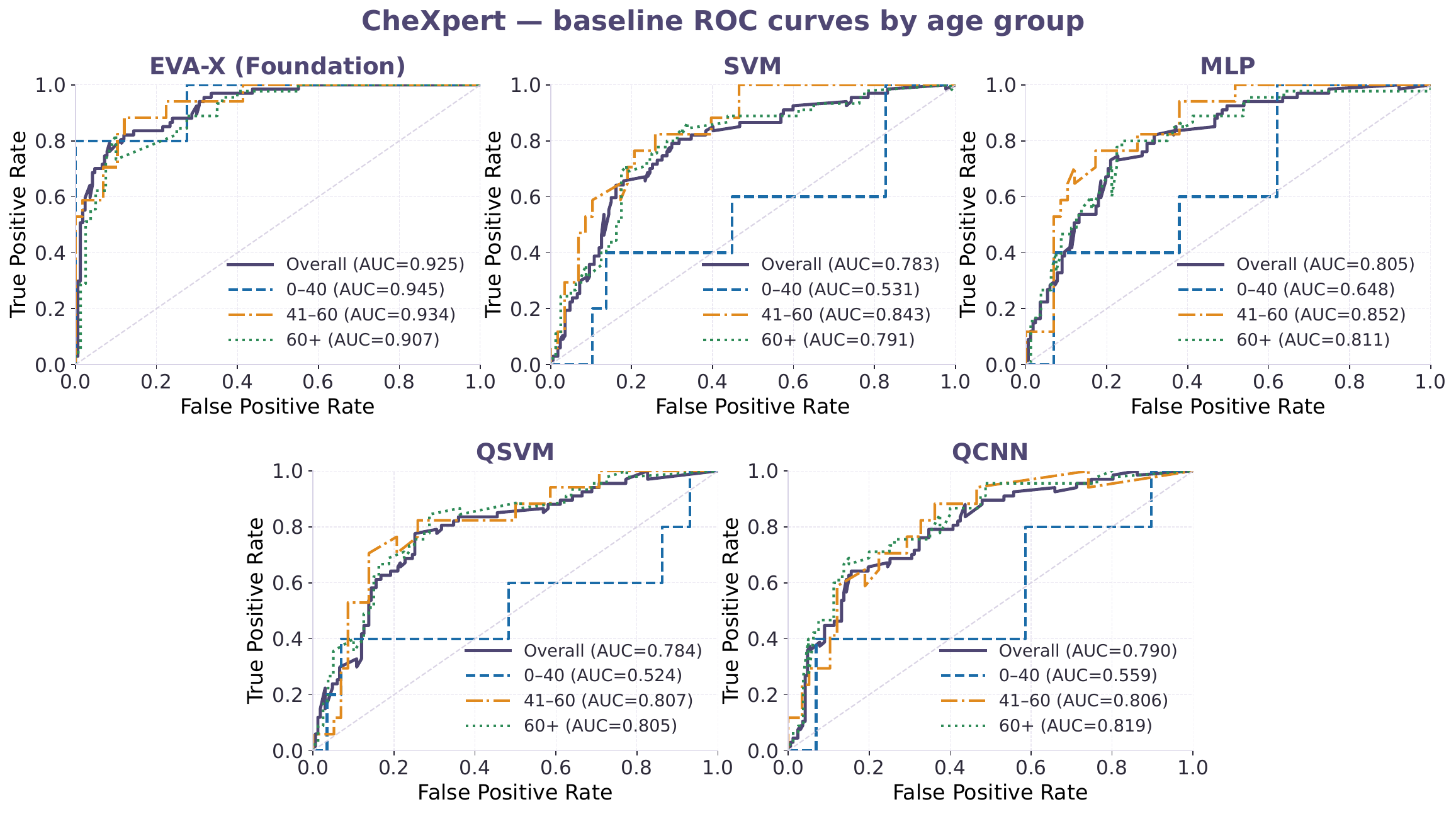}
    \caption{
    \textbf{Baseline ROC curves on CheXpert by age group.}
    Each panel shows the overall ROC curve and subgroup-specific curves for the 0--40, 41--60, and 60+ age groups. The downstream classifiers show weaker discrimination for the youngest subgroup, highlighting the importance of subgroup-aware evaluation.
    }
    \label{fig:roc_chexpert}
\end{figure}

Two caveats apply to the subgroup values reported in this section and in Figure~\ref{fig:roc_chexpert}. First, subgroup AUCs are computed on a fraction of an already modest held-out test set, so their effective sample sizes are considerably smaller than that of the global estimate and they are correspondingly less stable~\cite{Hanley1982ROC}; the low values observed for the 0--40 age group in CheXpert, in particular, rest on the smallest of the three subgroups. Second, no resampling-based confidence intervals were computed for these estimates~\cite{Efron1993Bootstrap}. The subgroup values should therefore be read as indicating the presence and approximate direction of disparity, rather than as precise measurements of its magnitude.

Figure~\ref{fig:roc_chexpert} shows the corresponding baseline ROC curves for CheXpert. The EVA-X foundation model presented high and relatively stable subgroup AUCs, with values of 0.945 for the 0--40 group, 0.934 for the 41--60 group, and 0.907 for the 60+ group. This corresponds to a comparatively small subgroup difference of 0.038, suggesting that the foundation model maintained strong discrimination across age groups in this dataset.

In contrast, the downstream classifiers appeared to show a consistent reduction in performance for the youngest age group. For SVM, the AUC was 0.531 for the 0--40 group, compared with 0.840 for 41--60 and 0.791 for 60+. A similar pattern was observed for MLP, with AUCs of 0.648, 0.852, and 0.811 across the same age groups. The quantum models followed the same tendency: QSVM obtained 0.524 for 0--40, compared with 0.807 for 41--60 and 0.805 for 60+, while QCNN obtained 0.559, 0.806, and 0.819. Therefore, although the overall AUCs of the downstream classifiers were close to 0.78--0.81, their discrimination for the 0--40 subgroup may be substantially weaker.

It is worth noting that the 0--40 group is also the group with the lowest prevalence of pleural effusion (28.5\%, Figure~\ref{fig:class_distribution}). Part of the observed weakness may therefore reflect the smaller number of positive cases available for this subgroup, in training and in evaluation, rather than a property of the classifiers themselves~\cite{Larrazabal2020GenderImbalance,Zong2023MEDFAIR}. This does not diminish the clinical relevance of the finding---a model that discriminates poorly in younger patients is a model that discriminates poorly in younger patients, whatever the cause---but it does bear on the choice of remedy, since a disparity driven by subgroup sample size calls for a data-level intervention rather than a representational one.

Together, Figures~\ref{fig:roc_ddi} and~\ref{fig:roc_chexpert} suggest that subgroup-aware fairness auditing is necessary in medical QML workflows. In DDI, the largest subgroup differences were mainly associated with variation across Fitzpatrick groups, while in CheXpert the main disparity appeared in the 0--40 age group. These findings indicate that aggregate ROC curves and global AUC values may obscure clinically relevant subgroup behaviour, including in quantum classifiers.

\FloatBarrier

\subsection{AUC-based performance--fairness trade-off}\label{subsec:auc_tradeoff_results}

\begin{figure}[htbp]
    \centering
    \includegraphics[width=\textwidth]{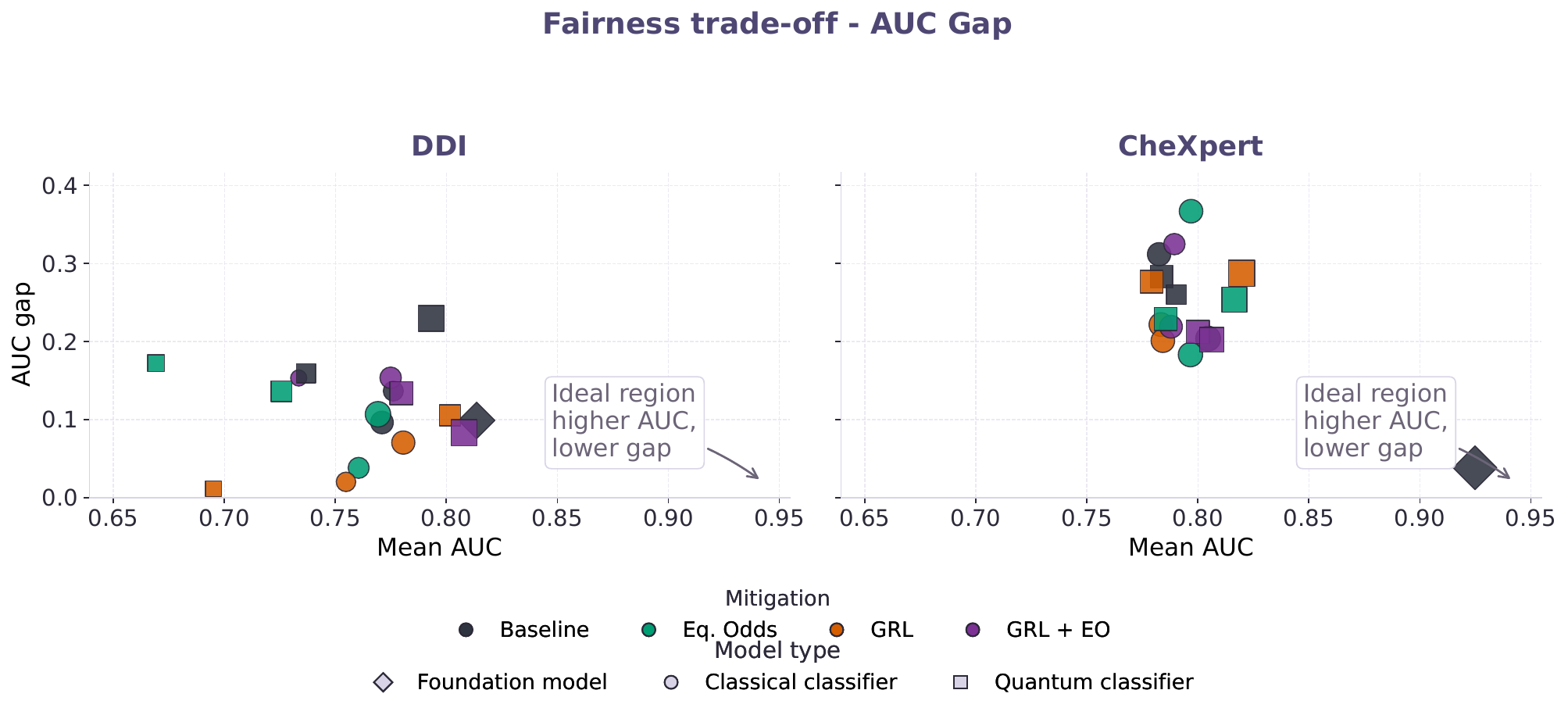}
    \caption{
    \textbf{Performance--fairness trade-off across models and mitigation strategies.}
    Each point represents a model and mitigation setting. The horizontal axis reports mean AUC, while the vertical axis reports AUC gap across sensitive groups. Colour indicates the mitigation strategy, marker shape indicates model type, and bubble size represents balanced accuracy.
    Downward movement alone does not imply clinical benefit; see Section~\ref{subsec:non_inferiority_results} for the joint criterion.
    }
    \label{fig:performance_fairness_tradeoff_auc}
\end{figure}

Figure~\ref{fig:performance_fairness_tradeoff_auc} summarises the relationship between predictive performance and subgroup disparity using AUC gap. The horizontal axis shows mean AUC, while the vertical axis shows AUC gap. Therefore, configurations located further to the right and lower on the plot may represent more favourable performance--fairness trade-offs, combining stronger discrimination with smaller differences across sensitive groups.

This reading requires one qualification. Position on the vertical axis alone does not establish that a configuration is preferable, because a low AUC gap can be reached either by improving the worst-performing group or by degrading the best-performing one~\cite{Mittelstadt2024LevellingDown}, and the plot does not distinguish the two. The horizontal axis must therefore be read jointly with the vertical one, and specifically relative to each model's own unmitigated baseline rather than in absolute terms. The acceptability region defined by Eq.~\ref{eq:non_inferiority}---configurations that move downwards (lower disparity) without moving left by more than $\varepsilon = 0.02$ relative to their baseline---is applied systematically in Section~\ref{subsec:non_inferiority_results}.

% NOTA PARA OS AUTORES: se houver tempo, regenerar trade_off.pdf com uma faixa
% vertical sombreada marcando AUC >= AUC_baseline - 0.02 para cada modelo,
% tornando a região de aceitabilidade visualmente explícita no próprio gráfico.

For DDI, the results suggest a trade-off between predictive performance and subgroup disparity. Some mitigation settings presented very low AUC gaps, such as QCNN with GRL and SVM with GRL, but these configurations were positioned further to the left in the plot, indicating lower mean AUC. Thus, although these settings may reduce subgroup disparity, they may not provide the most favourable overall trade-off when predictive performance is also considered. Section~\ref{subsec:non_inferiority_results} shows that this informal observation becomes a clear-cut exclusion once the non-inferiority margin is applied: QCNN with GRL attains the smallest AUC gap in the entire DDI experiment ($0.011$) while losing $0.042$ AUC points relative to its own baseline, and is therefore not an acceptable configuration under Eq.~\ref{eq:non_inferiority}.

A particularly relevant DDI result is observed for QSVM with the combined GRL+EO strategy. This configuration presented a mean AUC of 0.808 and an AUC gap of 0.083, placing it close to the PanDerm foundation-model reference in the trade-off space. PanDerm presented a mean AUC of 0.814 with an AUC gap of 0.099, meaning that QSVM with GRL+EO reached a similar predictive-performance level while showing a slightly smaller subgroup AUC gap. This suggests that, in this dataset and under the evaluated protocol, a quantum classifier combined with fairness-aware training may approach the state-of-the-art reference model while maintaining competitive subgroup fairness. Its disparity reduction is also not purchased at the cost of discrimination: relative to the unmitigated QSVM baseline it reduces the AUC gap by $0.146$ while \emph{increasing} global AUC by $0.015$, and it therefore satisfies Eq.~\ref{eq:non_inferiority} comfortably, with the largest admissible disparity reduction among all DDI configurations (Table~\ref{tab:non_inferiority}).

For CheXpert, the EVA-X foundation model occupied the most favourable region of the plot, with high mean AUC and low AUC gap. The downstream classifiers, including both classical and quantum models, were clustered around lower mean AUC values and higher AUC gaps. Among the non-foundation models, configurations such as MLP with Equalized Odds and QSVM with GRL+EO showed comparatively lower AUC gaps, but they remained clearly separated from the EVA-X reference. QCNN with GRL presented the highest global predictive point estimate among the non-foundation CheXpert models, but its AUC gap was not among the lowest, suggesting a less favourable fairness trade-off. Indeed, relative to its own baseline this configuration \emph{increases} the AUC gap by $0.027$ while gaining $0.030$ AUC points, and it therefore fails the disparity-reduction condition of Eq.~\ref{eq:non_inferiority} despite being the strongest non-foundation model on aggregate performance. This is the mirror image of the DDI case discussed above, and it illustrates that the two conditions of the criterion are genuinely independent.

Overall, Figure~\ref{fig:performance_fairness_tradeoff_auc} suggests that the effect of fairness mitigation is model (and dataset) dependent. Equalized Odds, GRL, and GRL+EO did not uniformly improve the trade-off across all settings. In DDI, the combined GRL+EO strategy appeared particularly relevant for QSVM, placing the quantum model close to the foundation-model reference. In CheXpert, however, the foundation model remained substantially better positioned than the downstream classifiers. These observations support the view that fairness interventions should be evaluated empirically in QML workflows, rather than assumed to consistently improve subgroup behaviour.

\FloatBarrier

\subsection{AUC gap analysis}\label{subsec:auc_gap_results}

\begin{figure}[htbp]
    \centering
    \includegraphics[width=0.88\textwidth]{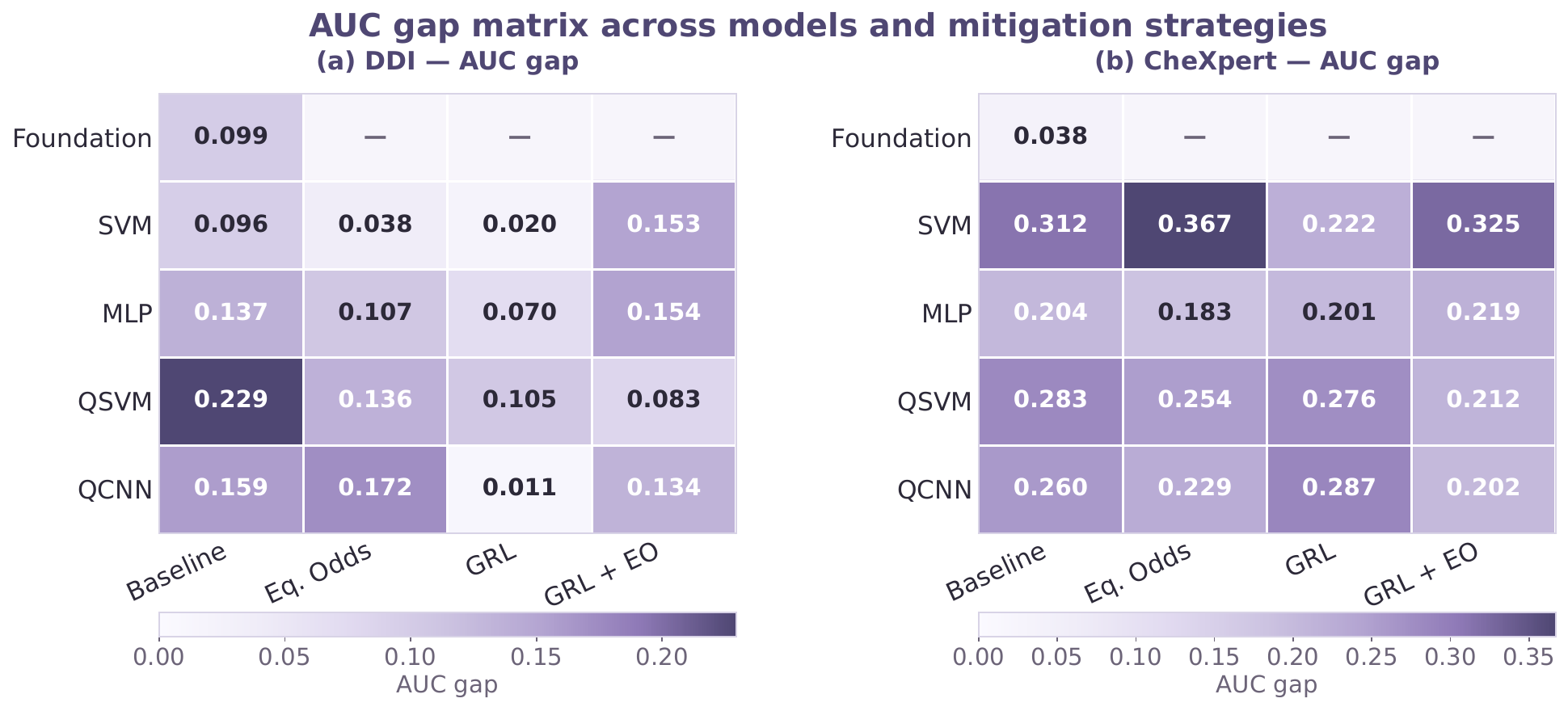}
    \caption{
    \textbf{AUC gap summary across models and mitigation strategies.}
    The heatmap reports the AUC gap between the best-performing and worst-performing sensitive groups. Lower values indicate smaller subgroup disparity. The two panels compare DDI and CheXpert under baseline and fairness mitigation settings.
    The heatmap reports disparity only; it should be read together with Table~\ref{tab:non_inferiority}, which reports the accompanying change in global discrimination.
    }
    \label{fig:auc_gap_heatmap}
\end{figure}

Figure~\ref{fig:auc_gap_heatmap} provides a direct comparison of AUC gaps across datasets, models, and mitigation strategies. Since lower AUC gap indicates smaller subgroup disparity in discriminative performance, this figure serves as the main visual summary of threshold-independent fairness behaviour.

On DDI, the smallest AUC gaps were observed for QCNN with GRL and SVM with GRL, with AUC gaps of 0.011 and 0.020, respectively. SVM with Equalized Odds also presented a low AUC gap of 0.038. These values suggest that, for DDI, some mitigation strategies may reduce subgroup ranking disparities. However, this reduction did not necessarily coincide with the highest global predictive performance. For example, QSVM with GRL+EO presented the highest global non-foundation point estimates in Table~\ref{tab:performance_ddi}, but its AUC gap was 0.083, which was not the lowest disparity value.

The converse relationship is more consequential and deserves to be stated explicitly. The two smallest disparities in the DDI panel are attained by configurations that also lose discrimination relative to their own baselines: QCNN with GRL falls from an AUC of $0.737$ to $0.695$, and SVM with GRL from $0.771$ to $0.755$. In the first case the loss exceeds the pre-specified margin $\varepsilon$. A reader comparing only the heatmap cells would rank QCNN with GRL as the fairest configuration in the entire DDI experiment, whereas the joint criterion classifies it as unacceptable, because its near-zero disparity is obtained in part by degrading the group that the baseline model served best. This is precisely the failure mode that motivates Eq.~\ref{eq:non_inferiority}~\cite{Mittelstadt2024LevellingDown,Zietlow2022LevelingDown}, and it is observed here in the study's own results rather than as a hypothetical concern.

On CheXpert, the AUC gap values were generally larger for the downstream classical and quantum classifiers. EVA-X presented the lowest AUC gap, equal to 0.038. Among the non-foundation models, the lowest AUC gaps were observed for MLP with Equalized Odds, QCNN with GRL+EO, and QSVM with GRL+EO, with values of 0.183, 0.202, and 0.212, respectively. Nevertheless, several mitigation strategies appeared to preserve or increase the AUC gap, particularly for SVM. This reinforces that mitigation effects may be model- and dataset-dependent.

\FloatBarrier

\subsection{Demographic parity gap analysis}\label{subsec:dp_gap_results}

Although AUC gap was used as the main subgroup-disparity metric, demographic parity gap was also analysed as a complementary fairness indicator. Unlike AUC gap, which measures group-wise discriminative performance, demographic parity gap measures differences in positive prediction rates across sensitive groups. Therefore, it should be interpreted carefully in medical tasks, because true disease prevalence may differ across age or skin type groups.

\begin{figure}[htbp]
    \centering
    \includegraphics[width=\textwidth]{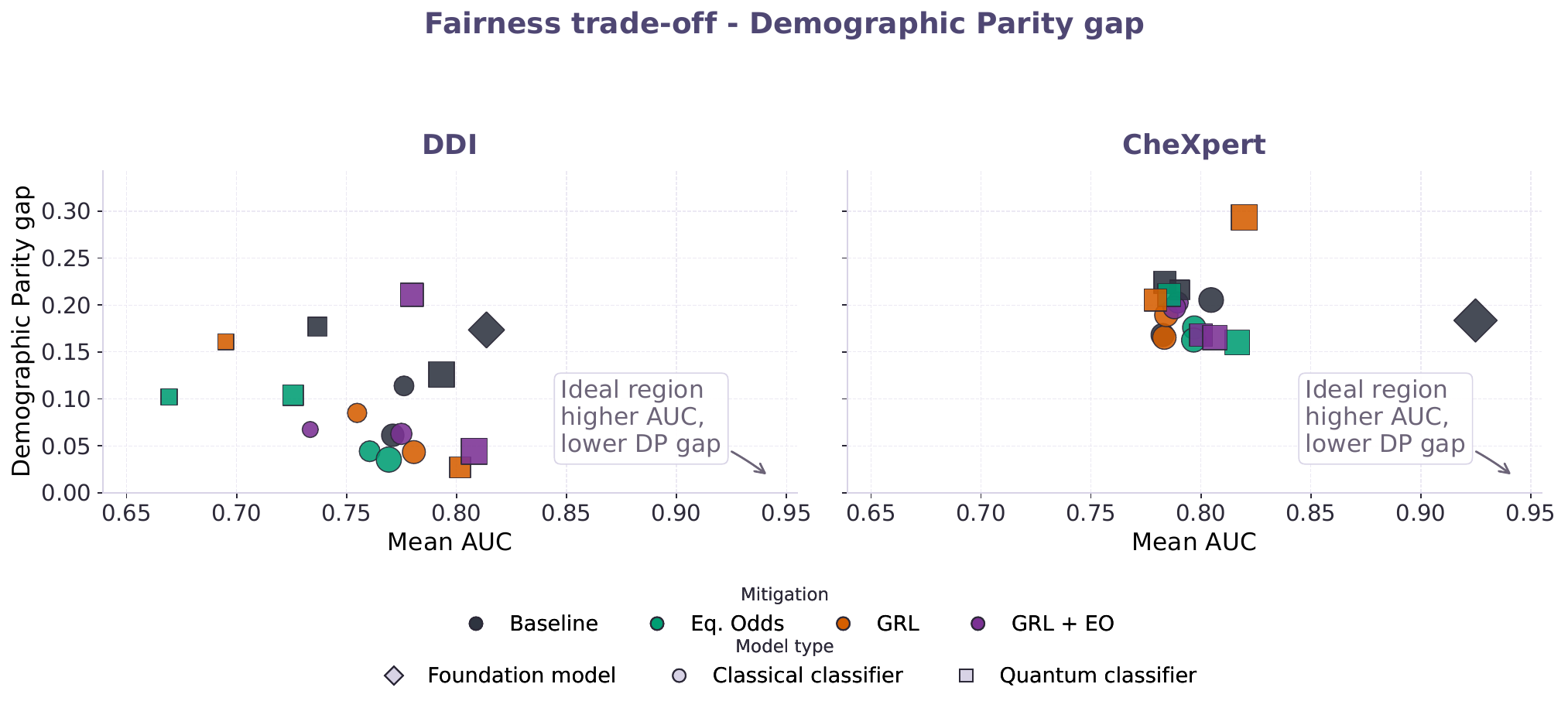}
    \caption{
    \textbf{Performance--fairness trade-off using demographic parity gap.}
    Each point represents a model and mitigation setting. The horizontal axis reports mean AUC, while the vertical axis reports demographic parity gap across sensitive groups. Colour indicates the mitigation strategy, marker shape indicates model type, and bubble size represents balanced accuracy.
    }
    \label{fig:performance_fairness_tradeoff_dp}
\end{figure}

Figure~\ref{fig:performance_fairness_tradeoff_dp} shows the performance--fairness trade-off using demographic parity gap on the vertical axis. The horizontal axis again reports mean AUC. Configurations located further to the right and lower on the plot may combine stronger predictive performance with smaller differences in positive prediction rates across groups.

For DDI, demographic parity gaps were generally smaller and more dispersed than the corresponding CheXpert values. Several non-foundation models occupied the lower region of the plot, which may indicate relatively small differences in positive prediction rates across Fitzpatrick groups. However, the foundation-model reference showed a higher demographic parity gap than some downstream classifiers, despite having strong global predictive performance. This illustrates that predictive performance and demographic parity may lead to different interpretations.

For CheXpert, the demographic parity gaps were more concentrated at higher values, which may reflect stronger differences in positive prediction rates across age groups. This behaviour is consistent with the class distribution shown earlier, where the prevalence of pleural effusion increased with age. Therefore, demographic parity results in CheXpert should not be interpreted as a direct measure of model error alone, but also as a metric that may reflect underlying prevalence differences between age groups. Following the distinction drawn in Section~\ref{subsubsec:chexpert}, a residual demographic parity gap in CheXpert is in fact the expected behaviour of a well-calibrated model on a population in which the condition is genuinely more common in older patients. Driving this gap to zero would require the model to predict pleural effusion at equal rates in populations where it occurs at unequal rates, which would introduce error rather than remove bias~\cite{Dwork2012Fairness,Chouldechova2017FairPrediction,Petersen2023EqualPerformance}. The CheXpert demographic parity values should accordingly be read as a descriptive property of the prediction distribution, and not as a quantity to be minimised.

\begin{figure}[htbp]
    \centering
    \includegraphics[width=0.92\textwidth]{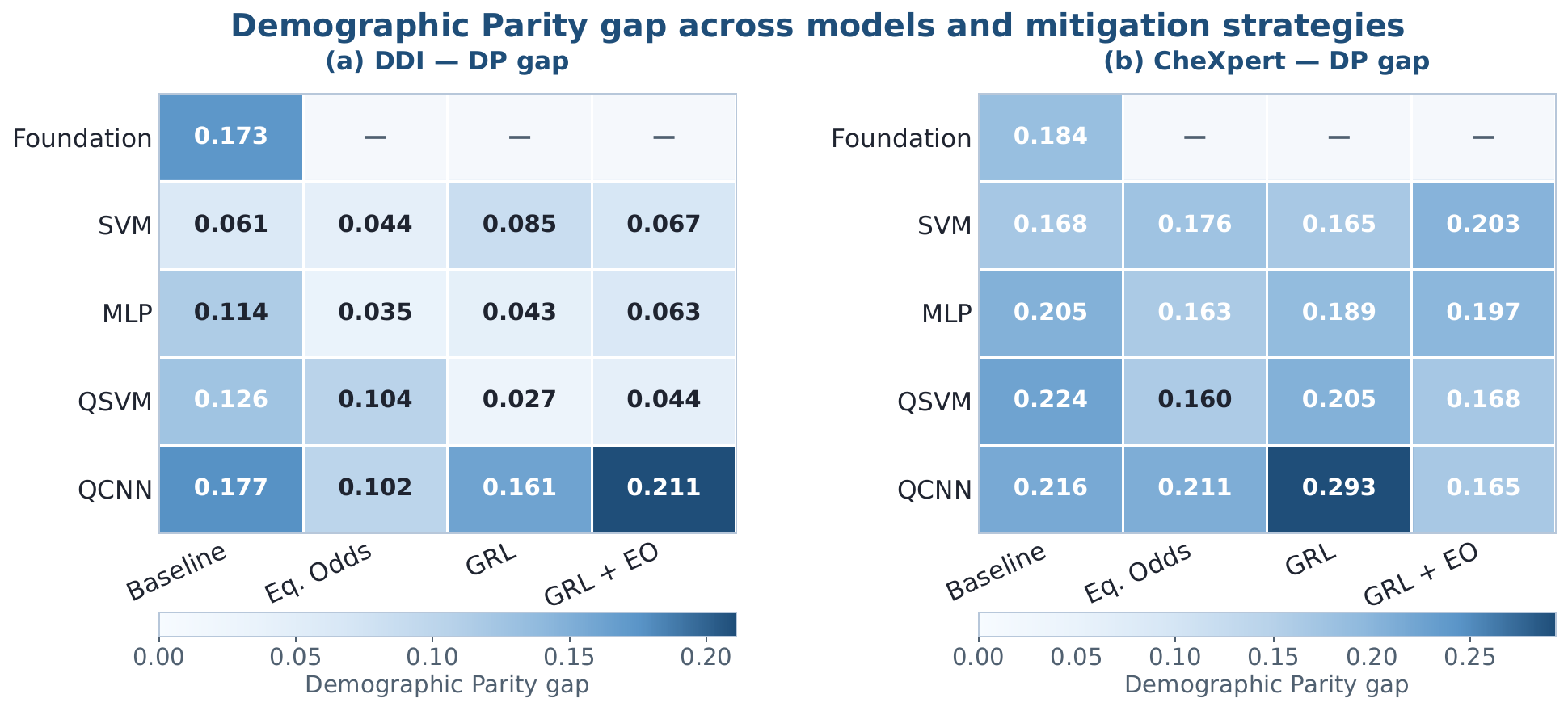}
    \caption{
    \textbf{Demographic parity gap summary across models and mitigation strategies.}
    The heatmap reports the demographic parity gap across sensitive groups. Lower values indicate smaller differences in positive prediction rates. The two panels compare DDI and CheXpert under baseline and fairness mitigation settings.
    }
    \label{fig:dp_gap_heatmap}
\end{figure}

Figure~\ref{fig:dp_gap_heatmap} provides a matrix summary of demographic parity gaps across models and mitigation strategies. This figure complements the AUC gap heatmap by showing how mitigation strategies may affect positive prediction-rate disparities.

In DDI, the lowest demographic parity gaps were observed for QSVM with GRL, MLP with Equalized Odds, and MLP with GRL, with values of 0.027, 0.035, and 0.043, respectively. These values suggest that some mitigation settings may reduce differences in positive prediction rates across Fitzpatrick groups. However, this did not always align with the AUC gap results. For instance, QCNN with GRL produced the lowest AUC gap in DDI but did not produce the lowest demographic parity gap.

In CheXpert, the lowest demographic parity gaps among downstream models were observed for QSVM with Equalized Odds, SVM with GRL, and QCNN with GRL+EO, with values of 0.160, 0.165, and 0.165, respectively. However, all CheXpert downstream models retained non-negligible demographic parity gaps. This may suggest that age-related differences in positive prediction rates remained present even after mitigation.

Overall, the demographic parity analysis supports the same general conclusion as the AUC gap analysis: fairness mitigation effects may not be uniform across datasets, model families, or fairness criteria. A mitigation strategy that improves one fairness metric may not improve another. For this reason, the results should be interpreted as a multi-metric fairness audit rather than as evidence that a single mitigation strategy is universally preferable.

\FloatBarrier

\subsection{Non-inferiority analysis of the mitigation strategies}\label{subsec:non_inferiority_results}

The analyses above report disparity and predictive performance separately. This section combines them by applying the criterion of Eq.~\ref{eq:non_inferiority} to every mitigated configuration, using each model family's own unmitigated baseline as the reference. Table~\ref{tab:non_inferiority} reports, for each configuration, the change in global discrimination $\delta_{\mathrm{AUC}}$, the change in subgroup disparity $\delta_{\Delta}$, and whether the configuration satisfies both conditions with $\varepsilon = 0.02$.

\begin{table}[htbp]
\centering
\scriptsize
\setlength{\tabcolsep}{6pt}
\renewcommand{\arraystretch}{1.15}
\caption{Joint assessment of disparity reduction and non-inferiority for each mitigated configuration, relative to the unmitigated baseline of the same model family.
$\delta_{\mathrm{AUC}}$ is the change in global AUC (Eq.~\ref{eq:delta_auc_global}) and $\delta_{\Delta}$ is the change in subgroup AUC gap (Eq.~\ref{eq:delta_gap}); negative $\delta_{\Delta}$ indicates reduced disparity.
A configuration is marked ``Yes'' only when it reduces disparity ($\delta_{\Delta} < 0$) \emph{and} does not lose more than $\varepsilon = 0.02$ AUC points ($\delta_{\mathrm{AUC}} \geq -0.02$).
Values are derived from Tables~\ref{tab:performance_chexpert} and~\ref{tab:performance_ddi} and from Figure~\ref{fig:auc_gap_heatmap}. Bold marks the largest admissible disparity reduction in each dataset.}
\label{tab:non_inferiority}
\begin{tabular}{llccc}
\toprule
\textbf{Model} & \textbf{Mitigation} & $\bm{\delta_{\mathrm{AUC}}}$ & $\bm{\delta_{\Delta}}$ & \textbf{Acceptable} \\
\midrule
\multicolumn{5}{c}{\textit{DDI (sensitive attribute: Fitzpatrick skin type)}} \\
\midrule
SVM  & Eq. Odds & $-0.010$ & $-0.058$ & Yes \\
SVM  & GRL      & $-0.016$ & $-0.076$ & Yes \\
SVM  & GRL+EO   & $-0.038$ & $+0.057$ & No \\
MLP  & Eq. Odds & $-0.007$ & $-0.030$ & Yes \\
MLP  & GRL      & $+0.005$ & $-0.067$ & Yes \\
MLP  & GRL+EO   & $-0.001$ & $+0.017$ & No \\
QSVM & Eq. Odds & $-0.067$ & $-0.093$ & No \\
QSVM & GRL      & $+0.009$ & $-0.124$ & Yes \\
QSVM & GRL+EO   & $+0.015$ & $\mathbf{-0.146}$ & \textbf{Yes} \\
QCNN & Eq. Odds & $-0.068$ & $+0.013$ & No \\
QCNN & GRL      & $-0.042$ & $-0.148$ & No \\
QCNN & GRL+EO   & $+0.043$ & $-0.025$ & Yes \\
\midrule
\multicolumn{5}{c}{\textit{CheXpert (sensitive attribute: age group)}} \\
\midrule
SVM  & Eq. Odds & $+0.014$ & $+0.055$ & No \\
SVM  & GRL      & $+0.000$ & $\mathbf{-0.090}$ & \textbf{Yes} \\
SVM  & GRL+EO   & $+0.006$ & $+0.013$ & No \\
MLP  & Eq. Odds & $-0.008$ & $-0.021$ & Yes \\
MLP  & GRL      & $-0.021$ & $-0.003$ & No \\
MLP  & GRL+EO   & $-0.017$ & $+0.015$ & No \\
QSVM & Eq. Odds & $+0.033$ & $-0.029$ & Yes \\
QSVM & GRL      & $-0.005$ & $-0.007$ & Yes \\
QSVM & GRL+EO   & $+0.016$ & $-0.071$ & Yes \\
QCNN & Eq. Odds & $-0.004$ & $-0.031$ & Yes \\
QCNN & GRL      & $+0.030$ & $+0.027$ & No \\
QCNN & GRL+EO   & $+0.016$ & $-0.058$ & Yes \\
\bottomrule
\end{tabular}

\vspace{0.4em}
\begin{minipage}{0.66\textwidth}
\footnotesize
\textit{Note:} The criterion constrains global discrimination only, because subgroup-level AUCs were computed for the baseline configurations. A stronger per-subgroup formulation, in which no individual sensitive group may deteriorate by more than $\varepsilon$, would require subgroup-level evaluation of every mitigated configuration and is identified as future work.
\end{minipage}
\end{table}

Three observations follow from Table~\ref{tab:non_inferiority}, together with an important caveat.

First, disparity reduction and predictive non-inferiority are genuinely dissociable, and the dissociation occurs in both directions within this study. On DDI, QCNN with GRL produces the single largest disparity reduction in the entire experiment ($\delta_{\Delta} = -0.148$, reaching an absolute gap of $0.011$) but loses $0.042$ AUC points and is therefore excluded. QSVM with Equalized Odds shows the same pattern more sharply, combining a substantial disparity reduction with a loss of $0.067$ AUC points. Conversely, on CheXpert, QCNN with GRL is the strongest non-foundation model on aggregate performance yet increases disparity, and is excluded for the opposite reason. Ranking mitigation strategies by disparity alone would have promoted exactly those configurations that a clinically oriented reading should reject~\cite{Mittelstadt2024LevellingDown,Petersen2023EqualPerformance}.

Second, five configurations satisfy the criterion on both datasets: SVM with GRL, MLP with Equalized Odds, QSVM with GRL, QSVM with GRL+EO, and QCNN with GRL+EO. Among them, QSVM with GRL+EO combines the largest admissible disparity reduction on DDI ($\delta_{\Delta} = -0.146$) with the largest admissible reduction among the quantum models on CheXpert ($\delta_{\Delta} = -0.071$), while improving global AUC in both cases ($+0.015$ and $+0.016$). It is not, however, the strongest configuration on every dataset: on CheXpert, the largest admissible disparity reduction was obtained by the classical SVM with GRL ($\delta_{\Delta} = -0.090$, with unchanged global AUC). The favourable positioning of QSVM with GRL+EO noted in Section~\ref{subsec:auc_tradeoff_results} therefore survives the stricter reading, in the sense that this configuration did not obtain its fairness profile by sacrificing discrimination, but the benefit of mitigation is neither exclusive to the quantum models nor uniform across datasets.

Third, the criterion is more permissive on CheXpert than the raw disparity values might suggest, and this should be interpreted with care. CheXpert AUC gaps remain large in absolute terms for every downstream configuration (between $0.183$ and $0.367$), so a configuration marked ``Yes'' in the lower panel of Table~\ref{tab:non_inferiority} has reduced disparity relative to its own baseline without losing discrimination, but has not achieved a small disparity. Non-inferiority is a relative criterion and does not certify that the residual disparity is clinically acceptable~\cite{Piaggio2012CONSORT}. In CheXpert, where the youngest age group shows baseline AUCs close to chance for several downstream models (Section~\ref{subsec:roc_results}), none of the mitigation strategies evaluated here brought subgroup performance to a level that would support responsible deployment, irrespective of how the disparity changed.

Finally, it must be emphasised that Eq.~\ref{eq:non_inferiority} constrains the \emph{global} AUC and not the AUC of each subgroup. A configuration could in principle satisfy the criterion while still degrading one subgroup, provided the loss is offset elsewhere. Distinguishing the two mechanisms directly---that is, establishing for each mitigated configuration whether the worst-performing group improved or the best-performing group deteriorated---requires subgroup-level AUC values for all mitigated settings, which were not computed in this study. This is the most important analytical extension identified for future work, and it is discussed further in Section~\ref{sec:limitations_future_work}.

\FloatBarrier

\section{Limitations and Future Work}\label{sec:limitations_future_work}

This study should be interpreted in light of several limitations. First, all quantum models were evaluated through classical simulation using PennyLane. Although this approach allows controlled comparison between classical and quantum classifiers, it does not capture all constraints associated with execution on real quantum hardware, including device noise, limited connectivity, finite sampling, calibration drift, and queue-dependent execution conditions. Therefore, the present results should be understood as a simulation-based fairness audit of QML workflows, rather than as evidence of hardware-level clinical feasibility.

Second, the downstream classifiers relied on fixed image embeddings extracted with a ResNet-18 model pretrained on ImageNet. This design ensured that classical and quantum classifiers were compared under a common feature representation, but it also means that the observed fairness behaviour may partly depend on the classical embedding backbone. Biases introduced or preserved by the feature extractor can be inherited by both classical and quantum models. Future studies should evaluate alternative medical imaging encoders, domain-specific foundation models, and end-to-end representation-learning strategies to determine how feature extraction affects fairness in QML pipelines.

This limitation has a direct consequence for the attribution of the results, which should be stated plainly. The quantum models never operate on the medical images: all visual information is first processed by a classical ImageNet-pretrained ResNet-18 and then compressed to ten principal components before any quantum circuit is applied. A substantial and possibly dominant part of both the discriminative representation and any subgroup bias may therefore be established in the classical stage, upstream of the quantum component. The present experimental design does not permit a clean attribution of the observed fairness behaviour to the quantum component specifically, because differences between, for example, SVM and QSVM may arise from the dimensionality of the input space, from the PCA projection, from differences in effective capacity or regularisation, from the architecture, or from the quantum feature map. The decisive control---classical classifiers trained on exactly the same ten-dimensional PCA representation supplied to the quantum models, at matched capacity---was not performed and is the highest-priority extension of this work. In its absence, the appropriate interpretation of this study is as a fairness audit of hybrid classical--quantum pipelines, and not as evidence of a specifically quantum contribution to fairness behaviour~\cite{Schuld2021KernelMethods,SchuldKilloran2019FeatureSpaces}. Correspondingly, no claim of quantum advantage in predictive performance, in fairness, or in computational efficiency is made or supported here; on the contrary, the quadratic scaling of quantum-kernel construction discussed in Section~\ref{subsubsec:qsvm} represents a computational cost rather than a saving~\cite{Preskill2018NISQ}.

Third, the quantum models were restricted to ten-qubit simulations and compact projected feature vectors. This constraint was necessary to keep QSVM kernel evaluation and QCNN training computationally feasible, but it limits the dimensionality and expressiveness of the quantum models. Larger circuits, alternative feature maps, noise-aware ansätze, and hardware-efficient architectures should be explored in future work. In particular, future studies should investigate whether quantum-specific circuit design can reduce subgroup disparities without relying only on classical fairness interventions.

Fourth, the fairness analysis was conducted on two binary medical image classification tasks and two sensitive attributes: Fitzpatrick skin type in DDI and age group in CheXpert. Although these settings are clinically relevant, they do not cover the full range of demographic, acquisition, institutional, and clinical factors that may affect medical AI fairness. Future work should extend the audit to larger and more diverse datasets, additional sensitive attributes, intersectional groups, and external validation cohorts.

Fifth, the statistical treatment of uncertainty is limited. The deviations reported throughout correspond to variability across five random seeds and are not resampling-based confidence intervals for the metrics themselves; nor were paired statistical tests performed between configurations. This limitation is most consequential for the subgroup analyses, where effective sample sizes are considerably smaller than for the global estimates, and for comparisons between closely spaced configurations such as QSVM with GRL+EO and the PanDerm reference on DDI, whose point estimates differ by substantially less than their reported deviations. Such comparisons are stated in this work as indicating comparable performance and should not be read as claims of superiority or of formal equivalence~\cite{AltmanBland1995Absence}. Bootstrap confidence intervals for the global AUC, for each subgroup AUC, for the AUC gap, and for the baseline-to-mitigation difference, together with paired tests between models~\cite{DeLong1988}, would substantially strengthen the inferential basis of the audit and are a priority for future work~\cite{Efron1993Bootstrap}. Where per-sample test-set predictions have been retained, these quantities can be obtained by re-analysis without retraining.

Sixth, and related, the non-inferiority criterion introduced in Section~\ref{subsec:evaluation_metrics} constrains global discrimination only. The clinically more informative decomposition---establishing, for each mitigated configuration, whether the reduction in disparity arose from improvement of the worst-performing subgroup or from deterioration of the best-performing one---requires subgroup-level AUC values for all mitigated settings, which were computed here only for the baseline configurations. Reporting the full matrix of per-subgroup changes $\delta_{\mathrm{AUC},g} = \mathrm{AUC}_{g}^{(m)} - \mathrm{AUC}_{g}^{(b)}$, and applying the non-inferiority margin to every group individually rather than to the global estimate, is the single analytical extension that would most strengthen the clinical interpretation of these results~\cite{Petersen2023EqualPerformance,Zong2023MEDFAIR}.

Seventh, the evaluation does not characterise the clinical consequences of the errors that the models make. All metrics reported here are either threshold-independent or computed at a single calibrated operating point, and they weight false negatives and false positives equally. In the tasks considered, they are not equivalent: a missed malignant lesion in DDI and a missed pleural effusion in CheXpert carry consequences of a different nature from those of a false alarm, and a mitigation strategy that reduces the AUC gap while shifting the error profile towards false negatives could be clinically unacceptable despite appearing more equitable. A complete assessment would report subgroup-level sensitivity, specificity, positive predictive value, and negative predictive value at clinically motivated operating points---with particular attention to changes in sensitivity for malignant lesions after mitigation---and would complement these with a net-benefit analysis such as decision curve analysis, in order to establish whether the mitigation strategies deliver clinical benefit and not merely a reduction in statistical disparity~\cite{Vickers2006DCA,Vickers2016NetBenefit}. Positive and negative predictive values are prevalence-dependent and would therefore need to be reported per subgroup, given the prevalence differences documented in Section~\ref{subsubsec:chexpert}~\cite{AltmanBland1994PredictiveValues}.

Finally, a conceptual limitation applies to the interpretation of the mitigation strategies themselves. The adversarial component treats sensitive-attribute information in the learned representation as undesirable by construction, without distinguishing spurious demographic shortcuts from legitimate epidemiological signal~\cite{Geirhos2020Shortcut,Zhang2018Mitigating}. In CheXpert, where the prevalence of pleural effusion rises substantially with age, this distinction is material: enforcing invariance to age may remove clinically informative structure together with any shortcut. Demographic invariance and fairness are not synonymous in medical applications, and mitigation objectives designed for settings in which the sensitive attribute carries no legitimate predictive information should be transferred to clinical tasks with corresponding caution~\cite{Wachter2021FairnessBias,Petersen2023EqualPerformance}.

Finally, the mitigation strategies considered here were limited to differentiable Equalized Odds regularisation, GRL, and their combination. These methods did not uniformly improve all fairness metrics, reinforcing the need for multi-metric evaluation.

\FloatBarrier

\section{Conclusions}\label{sec:conclusions}

This work presented an empirical fairness audit of classical and quantum machine learning models for medical image classification. Two healthcare scenarios were evaluated: dermatological image classification on DDI, using Fitzpatrick skin type as the sensitive attribute, and chest X-ray classification on CheXpert-derived data, using age group as the sensitive attribute. The study compared classical baselines, quantum classifiers, and foundation-model references under baseline and fairness-mitigation settings, including differentiable Equalized Odds regularisation, Gradient Reversal Learning, and the combined GRL+EO strategy.

The main finding is that QML should not be treated as inherently fair. Although quantum classifiers achieved competitive predictive performance in some settings, their behaviour varied across sensitive groups. On DDI, QSVM with GRL+EO achieved the strongest non-foundation performance and was positioned close to the PanDerm foundation-model reference in the performance--fairness trade-off space. On CheXpert, QCNN with GRL achieved the strongest non-foundation predictive performance, although EVA-X remained the best-performing reference model. These results indicate that QML models may be competitive, but competitiveness in aggregate metrics does not remove the need for subgroup-aware evaluation.

The subgroup ROC curves, AUC gap analysis, and demographic parity analysis showed that models with similar global AUC values could present substantially different subgroup disparities. In CheXpert, the downstream classifiers showed weaker discrimination for the youngest age group, while in DDI, performance varied across Fitzpatrick skin type groups. These findings reinforce that global predictive performance is insufficient for assessing responsible use in medical applications.

The mitigation strategies produced model- and dataset-dependent effects. EO regularisation, GRL, and GRL+EO reduced disparity in some cases, but preserved or increased it in others. The comparison between AUC gap and demographic parity gap also showed that different fairness metrics may lead to different interpretations, especially when class prevalence differs across sensitive groups.

A further conclusion concerns how such disparity reductions should be read. Applying an explicit non-inferiority criterion to every mitigated configuration showed that reduced disparity and preserved discrimination are dissociable properties, and that the dissociation occurs in both directions within these results~\cite{Mittelstadt2024LevellingDown,Zietlow2022LevelingDown}. On DDI, the configuration achieving the smallest subgroup disparity in the entire experiment---QCNN with GRL, with an AUC gap of $0.011$---does so while losing $0.042$ AUC points relative to its own baseline, and is therefore not an acceptable configuration under the criterion adopted here. On CheXpert, the strongest non-foundation model on aggregate performance increases disparity rather than reducing it. Ranking mitigation strategies by a disparity metric alone would have promoted precisely these configurations. QSVM with GRL+EO, by contrast, satisfies the criterion on both datasets, produces the largest admissible disparity reduction on DDI and the largest among the quantum models on CheXpert, and improves global discrimination in both cases; the favourable positioning of this configuration therefore survives the stricter reading, although on CheXpert the classical SVM with GRL achieved a larger admissible reduction. At the same time, the criterion is relative, and satisfying it does not certify that residual disparity is clinically acceptable: on CheXpert, subgroup performance for the youngest age group remained far below a level that would support responsible deployment under every mitigation strategy evaluated.

The broader message is accordingly not that fairness interventions can reduce disparities, which these results confirm they sometimes can, but that a reduction in a fairness gap should not be presented as intrinsically beneficial. In medical applications, a fairness intervention warrants being regarded as beneficial only when the reduction in unjustified disparity is achieved without clinically meaningful loss of diagnostic capability and without increased harm to patients~\cite{Petersen2023EqualPerformance,Piaggio2012CONSORT}. Establishing the latter requires evidence that the present audit does not provide---per-subgroup performance changes relative to baseline, metrics at clinically relevant operating points, and a net-benefit analysis---and these are identified above as the principal extensions of this work. It should also be noted that demographic invariance is not synonymous with fairness in clinical settings, since a sensitive attribute such as age may carry legitimate epidemiological information about disease probability that an invariance-enforcing objective will remove alongside any spurious shortcut~\cite{Wachter2021FairnessBias,Geirhos2020Shortcut}.

To the best of our knowledge, this is one of the first empirical studies to apply GRL as a fairness-mitigation strategy in QML-based medical image classification, and also one of the first to evaluate a combined GRL+EO differentiable mitigation pipeline in QML workflows. Overall, the findings support a stronger and more cautious conclusion: QML models should be evaluated under the same subgroup-aware and responsible-AI protocols expected from classical medical AI systems. Quantum classifiers may offer different representation mechanisms, but they remain subject to data imbalance, subgroup variation, and fairness trade-offs.

Finally, the scope of the quantum claim should be stated explicitly. All quantum models were evaluated in classical simulation, and all of them operate on ten principal components extracted from a classical ImageNet-pretrained backbone rather than on the medical images themselves. Without a classical control trained on that same ten-dimensional representation at matched capacity, the contribution of the quantum component cannot be isolated from that of the classical feature extractor, the dimensionality reduction, or differences in model capacity and regularisation. This work therefore demonstrates that hybrid classical--quantum pipelines exhibit subgroup disparities and require subgroup-aware auditing, and it does not demonstrate a quantum advantage in predictive performance, in fairness behaviour, or in computational efficiency~\cite{Preskill2018NISQ}.

\FloatBarrier

\subsection*{Funding}
This study was funded by the Fundação de Amparo à Pesquisa do Estado de São Paulo (FAPESP) through Project Nos. 2025/19585-6 (G.S.F.), 2024/14934-0
(F.M.), 2023/04987-6 and 2024/00998-6 (F.F.), and by the Conselho Nacional de Desenvolvimento Científico e Tecnológico (CNPq) through Project No. 408884/2024-0 (F.F.).
%==============
% References
% =========================
\bibliographystyle{bst/sn-aps}
\bibliography{sn-bibliography}

\end{document}